\documentclass[letterpaper, 10 pt, journal, twoside]{ieeetran}  
\usepackage{amsmath} % assumes amsmath package installed
\usepackage{amssymb}  % assumes amsmath package installed
\usepackage{graphics} % for pdf, bitmapped graphics files
\usepackage{epsfig}
\usepackage[font={footnotesize}]{caption}
\usepackage{subcaption}
\usepackage{balance}
\usepackage[normalem]{ulem}
\usepackage{color}
\usepackage{soul}
\RequirePackage[colorlinks,citecolor=blue,urlcolor=blue]{hyperref}

\title{Action Anticipation: Reading the Intentions of Humans and Robots} %Use for final RAL version

\author{Nuno Duarte$^{1 \href{https://orcid.org/0000-0002-1396-6774} {\includegraphics[scale=0.025]{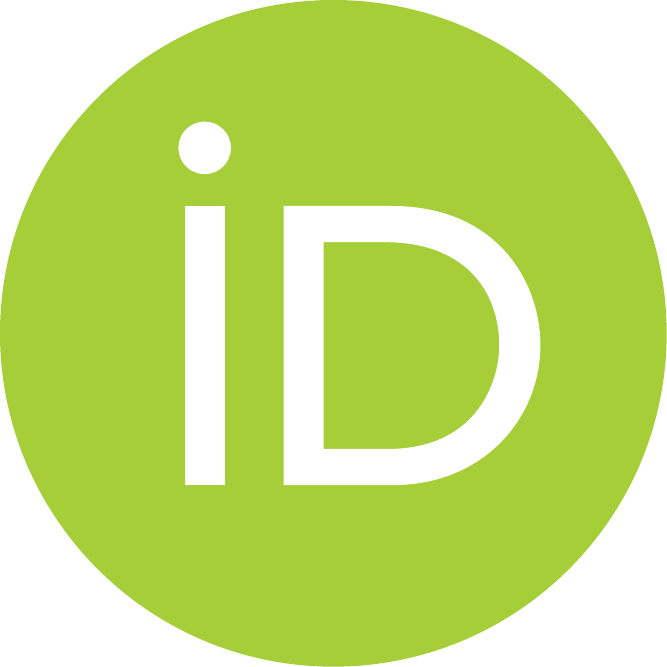}}}$, Mirko Rakovi\'{c}$^{1,2 \href{https://orcid.org/0000-0002-8818-2739}{\includegraphics[scale=0.025]{Images/orcid}}}$, Jovica Tasevski$^{2\href{https://orcid.org/0000-0003-2431-2642}{\includegraphics[scale=0.025]{Images/orcid}}}$, Moreno Coco$^{3 \href{https://orcid.org/0000-0002-2825-4200}{\includegraphics[scale=0.025]{Images/orcid}}}$, Aude Billard$^{4 \href{https://orcid.org/0000-0002-7076-8010}{\includegraphics[scale=0.025]{Images/orcid}}}$ and Jos\'{e} Santos-Victor$^{1 \href{https://orcid.org/0000-0002-9036-1728}{\includegraphics[scale=0.025]{Images/orcid}}}$ % <-this % stops a space
\thanks{Manuscript received: February, 23, 2018; Revised May, 30, 2018;
Accepted July, 11, 2018.}%Use only for final RAL version
\thanks{This paper was recommended for publication by Editor Dongheui Lee upon
evaluation of the Associate Editor and Reviewers' comments.}
\thanks{*This work was supported by EU H2020 project 752611 - ACTICIPATE, FCT project UID/EEA/50009/2013 and RBCog-Lab research infrastructure. We thank all colleagues that helped preparing and conducting the experiments, and all the people that participated in the human studies.}% <-this % stops a space
\thanks{$^{1}$N. Duarte,  M. Rakovi\'{c}  and J. Santos-Victor are with the Vislab, Institute for Systems and Robotics, Instituto Superior T\'{e}cnico, Universidade de Lisboa, Portugal,
	{\tt\small $\{$nferreiraduarte, rakovicm,  jasv$\}$@isr.tecnico.ulisboa.pt}}%
\thanks{$^{2}$J. Tasevski is with the Faculty of Technical Sciences, University of Novi Sad, Novi Sad, Serbia,
        {\tt\small tasevski@uns.ac.rs}}%
\thanks{$^{3}$Moreno Coco is with the Department of Psychology (Centre for Cognitive Ageing and Cognitive Epidemiology), University of Edinburgh, Scotland,
        {\tt\small moreno.coco@ed.ac.uk}}%
\thanks{$^{4}$Aude Billard is with the Learning Algorithms and Systems Laboratory, School of Engineering, EPFL, Lausanne, Switzerland {\tt\small aude.billard@epfl.ch}}%
\thanks{Digital Object Identifier (DOI): see top of this page.}
} 

\begin{document}

\maketitle

\begin{abstract}
Humans have the fascinating capacity of processing non-verbal visual cues to understand and anticipate the actions of other humans. This ``intention reading'' ability is underpinned by shared motor-repertoires and action-models, which we use to interpret the intentions of others as if they were our own.

We investigate how the different cues contribute to the legibility of human actions during interpersonal interactions. Our first contribution is a publicly available dataset with recordings of human body-motion and eye-gaze, acquired in an experimental scenario with an actor interacting with three subjects. From these data, we conducted a human study to analyse the importance of the different non-verbal cues for action perception. As our second contribution, we used the motion/gaze recordings to build a computational model describing the interaction between two persons. As a third contribution, we embedded this model in the controller of an iCub humanoid robot and conducted a second human study, in the same scenario with the robot as an actor, to validate the model's ``intention reading'' capability.

Our results show that it is possible to model (non-verbal) signals exchanged by humans during interaction, and how to incorporate such a mechanism in robotic systems with the twin goal of : (i) being able to ``read'' human action intentions, and (ii) acting in a way that is legible by humans. 

\end{abstract}

% Keywords appear just beneath the abstract. Use only for final RAL version.
%%%%%%%%%%%%%%%%%%%%%%%%%%%%%%%%%%%%%%%%%%%%%%%%%%%%%%%%%%%%%%%%%%%%%%%%%%%%%%%%

%\input{1_introduction.tex}
\section{Introduction} \label{sec:introduction}

\begin{figure}[t]
      \centering
      \includegraphics[width = 0.48\textwidth]{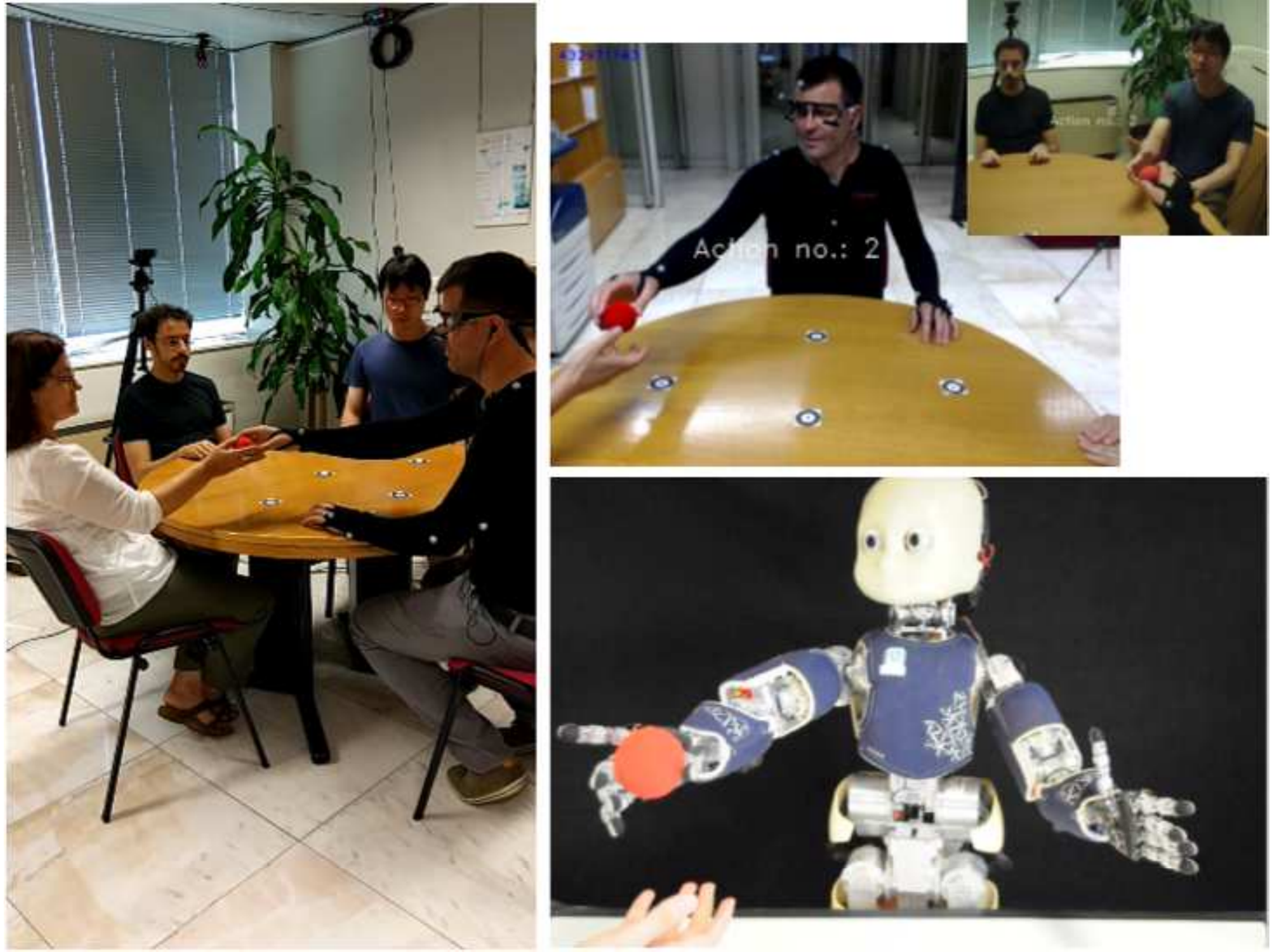}
      \caption{Human-Human Interaction: an experiment involving one actor (top-right) \textit{giving} and \textit{placing} objects and three subjects reading the intentions of the actor (left); Human-Robot Interaction: a robot performing the human-like action and subjects try to anticipate the robots' intention (bottom-right).} 
      \label{fig:HHItoHRI}
	\end{figure}

\IEEEPARstart{W}{hen} working in a shared space, humans interpret non-verbal cues such as eye gaze and body movements to understand the actions of their workmates. By inferring the actions of others, we can efficiently adapt our movements and appropriately coordinate the interaction (Fig. \ref{fig:HHItoHRI}).  According to Dragan et al~\cite{dragan2013legibility}, the intention of others can only be understood if and when the end-goal location becomes unambiguous to us. For that same reason, to improve human-robot interaction (HRI), robots should perform coordinated movements of all body parts, so that their actions and goals can be ``legible'' to humans.

Recent research in HRI has focused on studying the human behaviour 
\cite{erlhagen2006goal, huber2008human, marin2009interpersonal, sciutti2018humanizing}. Several papers, which we will discuss in more detail in Section \ref{sec:problem_identification}, have built bio-inspired controllers that facilitate human action understanding and interaction, and improve the communication with robots. However, they do not focus on the essential part of human interaction - the communication of intent - 
the central focus of our work.

We start by defining a scenario of human-human interaction (HHI), detailed in Section \ref{sec:humans_experiment}, to study non-verbal communication cues between humans, in a quantitative manner. 
The experiment consists of an actor performing goal-oriented actions in front of three humans sitting at a round table (Fig.\ref{fig:HHItoHRI}-left). The actor picks  up a ball placed in front of him and has to either (i) \textit{place} the ball on the table in front of one of the three persons or (ii) \textit{give} the ball to one of them (Fig.\ref{fig:HHItoHRI} - top right). Considering the two actions (placing/giving) and three spatial parametrizations (left/middle/right), the actor executes one out of six action-possibilities.
With this HHI experiment, we have built a dataset with the actor's 3D body movements and eye-gaze information during the interaction. Additionally, video recordings were taken during the entire experiment, and used to design a human study.

The videos of the actor are used to analyse three different cues: eye gaze, head orientation, and arm movement towards the goal position (Section \ref{sec:humans_cues}) of \textit{placing} and \textit{giving} actions. For this study, we prepared a gated experiment, using a set of video segments of increasing temporal duration, of each action performed by the actor. The video fractions are shown to the participants, and they are asked to predict the actor's intended action: \textit{giving} the ball to one of the persons or \textit{placing} the ball at one of three assigned markers on the table (6 possibilities in total).  Our results reveal that early eye-gaze shifts provide important information for the human subjects to anticipate the intention of the actor. Additionally, we observed significant of the eye-gaze behaviour between \textit{giving} and \textit{placing} actions, that seems to be governed by and attend to multiple goals.

The  recordings of the upper body and eye gaze motion are used to develop a computational model of the human actions (Section \ref{sec:modeling}). The arm movement was modelled with Gaussian Mixture Models (GMM), and Gaussian Mixture Regression (GMR) is used to generate the arm trajectory. The eye gaze behaviour  depends on the type of action. Before picking up the ball, the eye fixates the initial ball position. Then, for the \textit{placing} action, the eye gaze aims at the goal position (i.e. marker on the table). In case of the \textit{giving} action, the eye gaze switches between the face of the human and end-goal position (i.e. the handover location). 

The developed computational model is incorporated in a controller for the iCub humanoid robot, with the purpose of validating the model and investigating whether humans can ``read'' the robot actions in the same way they can read the actions of other humans. We have built a second human study for the same scenario using a robot actor. We recorded videos of the robot performing the same set of actions as the human actor. The video fractions of the robot-actor are then presented to another group of participants, who are asked to anticipate the robot's action intention (Section \ref{sec:robots_cues}). 

In Section \ref{sec:discussion} we discuss our experiments and results concerning the human perception of the robot's actions, in terms of readability. Our results show that we can model the non-verbal communication cues during human-human interaction and transfer that model to a robot executing \textit{placing} actions or \textit{giving} a ball to a human. Finally, we draw some conclusions and establish directions of future work.

\section{State of the Art}\label{sec:problem_identification}

Dragan et al~\cite{dragan2013legibility} discuss the aspects of predictability and legibility of arm movements. They define legible robot actions as copies of human actions but executed with exaggerated movements, and demonstrate that they can be understood sooner.
Instead, in our work, legibility is not achieved by exaggerating the arm movements, but by modelling the natural coordination of human eye, head, and arm movements. For that purpose, we conduct a quantitative analysis of the importance of the robot eye-gaze behaviour for the legibility of the robot's movements. We validate the model with a human study where subjects need to read the robot's intentions and select between  (\textit{placing}) or (\textit{giving}) actions with three spatial parametrizations.

Research in HRI and, more specifically, in human motion understanding \cite{activity2012forecasting, pfeiffer2016predicting, sciutti2015investigating} and modelling \cite{zhang2016rgb}, has relied on different existing datasets. Zhang et al. \cite{zhang2016rgb} present a survey on RGB-D based action recognition dataset. The CAD 120 dataset~\cite{koppula2016anticipating} includes a rich repertoire of human actions including the labels of the activities performed during those actions. Some of the existing datasets only provide information related to 3D body coordinates, while the few which include gaze tracking have the drawback of being limited to 1 or 2 tasks~\cite{gottwald2015good,admoni2014deliberate,fathi2011learning}. 
The first contribution of this paper is to provide a publicly-available dataset, that overcomes the shortcomings of existing datasets and contains synchronised and labelled video+gaze and body motion in a dyadic scenario of interaction.\footnote{The dataset of synchronised video, gaze fixations from Pupil eye tracker, and body motion from OptiTrack motion tracking system of \textit{placing} and \textit{giving} actions can be downloaded from: \href{http://vislab.isr.tecnico.ulisboa.pt/datasets/\#acticipate1}{dataset.ACTICIPATE.ral-2018}}. 
\href
This dataset has already been successfully used to develop a novel action anticipation algorithm, that integrates the cues from both gaze and body motion to provide faster and more accurate predictions of human's action \cite{schydlo2018}. 

Neurobiology provides extensive insight into the biological models of the human sensory-motor system. One group of neuroscientists have focused on investigating cortical structures such as the posterior parietal cortex, the premotor and the motor cortices \cite{goodale2011transforming}. Another stream of research has been directed on modelling the role of the cerebellum in the motor loop, movement generation and synchronisation of sensory-motor system \cite{ohyama2003cerebellum}. These findings are used in \cite{shukla2012coupled,lukic2015visuomotor9} to develop coupled dynamical systems framework for arm-hand and eye-arm-hand motion control for robots. The framework is focused on motor control coupling. Here, we extend our previous work, to the analysis of the interpersonal coordination of sensory-motor systems during interaction. Therefore our dataset of coordinated gaze and body movements during dyadic interactions is then used to build a bio-inspired model.

Authors in \cite{elsner2014infants} investigate the infants' perception during object-handover interactions. Those studies show that, in spite of their young age, the gaze behaviour is already modulated by the social interaction context. The work described in \cite{land1999roles} shows how the gaze behaviour encompasses multiple fixation points when the subject is engaged in complex tasks, such as tea-making. However, none of these works develops experiments with on-line tracking of the eye gaze, head orientation, and arm movements during an interpersonal interaction, with \textit{placing} or \textit{giving} actions in different spatial parametrizations.

Meng et al. \cite{zheng2015impacts} study human eye-gaze during interaction. They built an experiment where different types of gaze trajectories are examined in a human-robot scenario. However, their analysis is not based on a quantitative sensory system but, rather, by manually labelling at the subject's eyes in the video recordings. We propose using an eye-tracking system to record and assess the human gaze behaviour in those actions. Furthermore, they conclude that, for \textit{giving} actions, humans prefer when the robot fixates the person's face and then switches, i.e. looks, to the handover position, as opposed to just looking either the face or the handover position exclusively. This is a contextually based behaviour of the gaze that we intend to study using the eye-tracking system. 

The second set of limitations in \cite{zheng2015impacts}, \cite{PrezDArpino2015FastTP}, \cite{hrcollaborative2015julieshah} concerns the robot used in the experiments. Due to the limited number of degrees of freedom in the head of the robot, the eye gaze shifts are simulated with head rotation. In our work, we use an eye-tracking device to observe the actual gaze fixation points during the interaction independently from the head gaze as this provides better accuracy than just the head orientation \cite{palinko2016robot}. We use the iCub humanoid robot that has a human-like face where the eyes can independently move, and thus express a readable behaviour of eye-gaze and head-gaze. 

\section{Interaction Scenario}\label{sec:humans_experiment}

This section presents an interaction scenario for collecting: (i) videos of actor movements to study the contribution of different cues and timings on anticipation of actions and (ii) the motion of the eye-gaze and relevant body-parts of a human actor, to model the human movements. 

\subsection{Scenario Description}\label{subsec:exp_setup}

The scenario can be seen in Fig. \ref{fig:HHItoHRI}(left). For each trial, one actor executes a set of \textit{placing} or \textit{giving} actions directed towards one of the three (left/middle/right) subjects. The actor was instructed to act as normal as possible when performing those actions. The actor picks the object from the initial position and executes one of these 6 preselected action-configurations (2 actions and 3 spatial directions). 
\begin{itemize}
	\item \textbf{\textit{placing}} on the table to the actor's \textbf{left} ($P_{L}$),  \textbf{middle} ($P_{M}$), or \textbf{right} ($P_{R}$),
	\item \textbf{\textit{giving}} the ball to the person on actor's \textbf{left} ($G_{L}$),  \textbf{middle} ($G_{M}$), or \textbf{right} ($G_{R}$).
\end{itemize}
 
The actions to execute were instructed over an earpiece to the actor so that none of the other participants could know which would be performed next. The order of the actions is randomly selected to prevent the actor from adapting its posture prior to initiation. Every action begins with picking up the ball and ends with the actor placing the ball back to the initial position on the table.

\subsection{Hardware and Software Setup}\label{subsec:HW_SW_setup}

The actor movements were recorded with an OptiTrack motion capture (MoCap) system, consisting of 12 cameras all around the environment and a suit with 25 markers, placed on the upper torso, arms, and head, that is worn by the actor. The MoCap provides position and orientation data of all relevant body parts (head, torso, right-arm, left-arm). 

The eye gaze was recorded with the mobile, binocular Pupil-Labs eye tracker \cite{Kassner:2014:POS:2638728.2641695}, that allowed us to track the actor's fixation point. To track the head movements with the MoCap system, head markers were placed on the Pupil-Lab system. To record the scene, three video cameras are used to provide different viewing angles that will complement during the evaluation phase. The first camera provides the world-view perspective of the actor from the Pupil Labs eye tracking headset (top-right image in Fig. \ref{fig:HHItoHRI}, the small window on top). The second camera records the table top where the actions will take place. This one provides a continuous look at the table and all the actor's movements (Fig. \ref{fig:HHItoHRI} - top right). The third camera was located further from the scene, looking inwards, giving a proper reading of the subject's actions and an outlook of the experiment (Fig. \ref{fig:HHItoHRI} - left).

To collect all the sensory information, the OptiTrack's Motive and Pupil Lab's Pupil Capture software were used. Prior to recording, both sensors were calibrated. Custom software was developed to acquire the video of the actor's action. All the sensory data are captured on distributed machines and data are streamed through the Lab Streaming Layer \cite{kothe2018lab} for centralised storage and data synchronisation.

\subsection{Synchronization of Sensory Data}\label{subsec:Dataset}

A total of 120 trials are performed with action-configurations: $P_{L}$, $P_{M}$, $P_{R}$, $G_{L}$, $G_{M}$ and $G_{R}$ performed 20, 23, 17, 17, 19 and 24 times respectively. The binocular eye gaze tracking system recorded world camera video and eye gaze data at 60Hz, the motion capture system recorded the movements of the body at 120Hz, and video camera facing the actor, recorded video at 30Hz. The data from all sensing systems are streamed and collected at one place, with the timestamps of each sensing system as well as the internal clock information, that is used as a reference to synchronise all sensory flows.

\section{Reading the Intentions of Humans}\label{sec:humans_cues}

We conducted a human study to quantify how the different cues contribute to the ability to anticipate the actions of others, and how those cues are related to the spatial (left/middle/right) distribution. The study includes a questionnaire pertaining to the actions performed by an actor.

\subsection{Participants}\label{subsec:Participants}

The study involved 55 participants (40 male, and 15 female), age 31.9$\pm$13 (mean$\pm$SD). There were 13 teenagers and 6 people over 50 years of age. Approximately 62$\%$ were students, 27$\%$ were professors, 7$\%$ were researchers, and 4$\%$ were staff members, 3 subjects were left-handed. All subjects were naive with respect to the purpose of the research.

\subsection{Human Study}\label{subsec:Questionnaire} 

The subjects were presented with videos of an actor performing \textit{giving} or \textit{placing} actions in the different spatial directions, and were asked to reply to a questionnaire related to the action being executed. The questionnaire consists of 24 questions\footnote{A description of the human study can be seen at the following web address: \href{http://vislab.isr.tecnico.ulisboa.pt/wp-content/uploads/2018/07/acticipate1_questionnaire_description.pdf}{files.ACTICIPATE.ral-2018}}. Before the question is shown to the subject, they have to watch a short video of the actor performing one of the six possible actions (2 end-goal actions multiplied by 3 directional end-goal locations). Based on the video shown, the participant had to identify the actor's intended action. The videos were fractioned into four types according to the cues provided by the eye gaze shift, head gaze shift, and arm movement. This can be understood as a gated experiment in which fractions of video segments are shown to subjects beginning when the actor grabs the object and ending when:
\begin{itemize}
    \item there is a saccadic eye movement towards the goal - G
    \item 'G' plus the head rotates to the same goal - G+H
    \item 'G+H' plus the arm starts moving to the goal - G+H+A
    \item 'G+H+A' plus the arm finishes the trajectory to the goal - G+H+A+.
\end{itemize}

The last group of videos (G+H+A+) was used as a golden standard to remove outliers. Out of 24 questions, the first three were used to familiarise participants with the questionnaire and were discarded from the analysis. Out of the remaining 21 questions, five questions are from the G difficulty level, six are from G+H, six are from G+H+A, while four were used for detecting outliers. Twelve are for \textit{placing} and nine are for \textit{giving} actions, whereas seven belong to left, eighth to middle and six to right direction.

\subsection{Analysis}
\label{stats}

From Section \ref{subsec:Questionnaire} we reach 5 important conclusions, that we describe in the following paragraphs.

\begin{figure}[ht]
	\centering
		\begin{subfigure}{0.49\textwidth}
			\includegraphics[width = 0.99\textwidth]{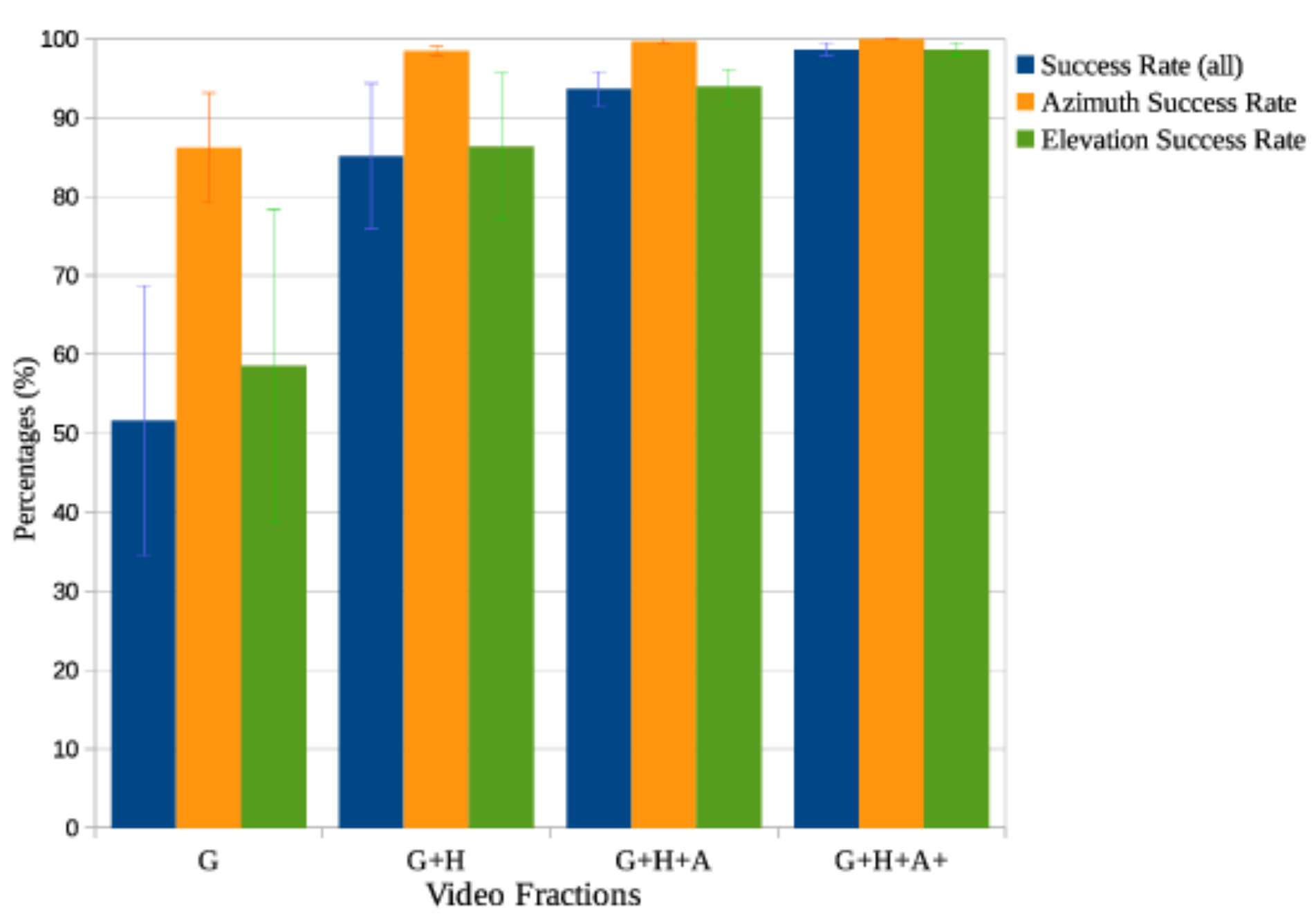}
			\caption{}\label{fig:overall_success}
		\end{subfigure}
        \hspace{2cm}
		\begin{subfigure}{0.49\textwidth}
			\includegraphics[width = 0.99\textwidth]{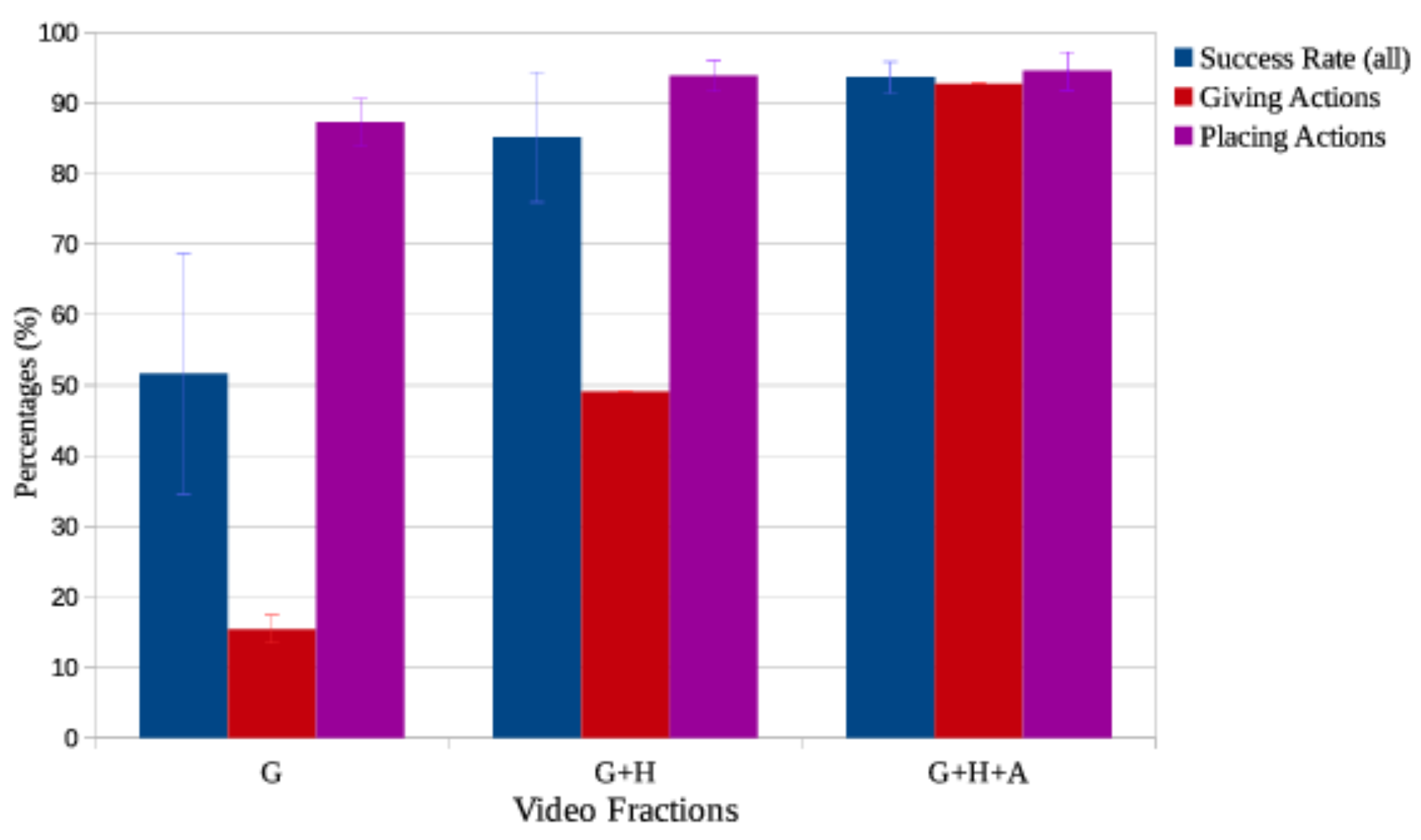}
			\caption{}\label{fig:placing_giving_success}
		\end{subfigure}		
	\caption{The success of the participants identifying the correct action a) overall success rate; success rate in identifying the direction of the action; b) success rate in identifying the \textit{giving} and \textit{placing} actions.}
	\label{fig:plots_human_human}
\end{figure}

The first conclusion is the most obvious and is shown in Fig. \ref{fig:plots_human_human}(a). The more temporal information is available to subjects, the better the decision is, the higher the success rate and the lower the variance. We validate this trend with a quantitative analysis, with a two-way ANOVA \cite{montgomery2010applied}, that shows a very significant correlation between the amount of information and the success rate, F(2,5560)=1396.76, p$<$0.0001. Gaze alone is responsible for a 50\% success rate of  (about 3 times the chance level of 1/6 = 16.7\%).  

The analysis is further refined by considering two variations: (i) how well can the subjects predict spatial orientation, irrespective of the \textit{giving} vs \textit{placing} action? and (ii) how can the subjects predict the action (\textit{giving}, or \textit{placing}) irrespective of the orientation (\textbf{left}, \textbf{middle}, or \textbf{right})?

Secondly, according to our results, the prediction of spatial orientation does not depend strongly on the amount of temporal information. The participants did not report significant difficulties to understand the gaze orientation from the 'G' videos when the actor was wearing the eye tracker, compared to a case where no glasses are used. Gaze alone is crucial for action understanding in the azimuth orientation, 85\% success (chance level of 33\%), then head information only increases around 15\%. Instead, action prediction depends strongly on the amount of temporal information. Surprisingly, subjects were only capable of understanding the action-type 60\% (chance level of 50\%) of the time for the first video fraction, but as more information was provided the success rate increased quite rapidly. To analyse in more detail the reason why, we refined this results in Fig. \ref{fig:plots_human_human}(b) to study two conditions: (i) \textit{giving} actions and (ii) \textit{placing} action.

Thirdly, we observe a significant interaction between the type of action and the amount of information available to the subject. This is confirmed by the two-way ANOVA, F(2,5560)=537.70, p$<$0.0001. For the \textit{placing} action we have a success rate of 85\% (chance level 50\%) with gaze alone. However, we observe that for the \textit{giving} action we get a success rate lower than chance level. Our fourth conclusion comes from the two-way ANOVA, confirming a significant importance between type of action and subjects' success rate, F(1,5560)=2306.78, p$<$0.0001, indicating a bias towards \textit{placing} in this HHI scenario.

These experiments clearly demonstrate, quantitatively, the importance of gaze in a dyadic action. In HHI, eye gaze information provides the necessary information to predict the intention of the other subject. For \textit{giving} actions this is not the case, but we believe that the experimental setup geometry introduces a unintentional bias towards the action that requires the least energy, \textit{placing} the object on the table. We evaluate the bias towards \textit{placing} by showing additional videos segmented before any non-verbal cue (smaller than 'G' video fraction) and the results show that in the case of \textit{placing} vs \textit{giving}, the majority of people picked \textit{placing}, proving a significant preconception in this HHI scenario. Our final conclusion is our cornerstone of this paper. This analysis shows that human eye-gaze provides key information to read the action correctly, and justifies the need to include human-like, eye-gaze control, in order to improve action-legibility and anticipation as required for efficient human-robot interaction.

% %\input{5_modeling_motion.tex}
\section{Modeling Human Motion}\label{sec:modeling}

This section begins by explaining the modelling of the arm motion and then proceeds by analysing the eye movements.

We use a Gaussian Mixture Model (GMM)~\cite{Calinon07} to model the trajectories of the arm movement in a probabilistic framework. The motion is represented as a state variable ${\{\xi_j\}}_{j=1}^{N} \in \mathbb{R}^{3}$, where N is the total number of arm trajectories for all actions, and $\xi_j$ are the Cartesian coordinates of the hand for \textit{giving} or \textit{placing} actions. The GMM defines a joint probability distribution function over the set of data from demonstrated trajectories as a mixture of \textit{k} Gaussian distributions each one described by the prior probability, the mean value and the covariance matrix. 
\begin{align}
    \nonumber p(k) & = \pi_k \\
    p(\xi_j|k) & = \mathcal{N}(\xi_j; \mu_k, \Sigma_k ) \\
    \nonumber & = \frac{1}{\sqrt{(2\pi)^D |\Sigma_k|}} e^{-\frac{1}{2}((\xi_j-\mu_j)^T\Sigma_k^{-1}(\xi_j-\mu_j))}
\end{align}
where $\{\pi_k;\mu_k,\Sigma_k\}$ is the prior probability, mean value, and covariance, respectively, for each \textit{k} normal distribution.

\begin{figure}[h]
	\centering
	\framebox{\parbox{3.3in}{\includegraphics[width = 0.47\textwidth]{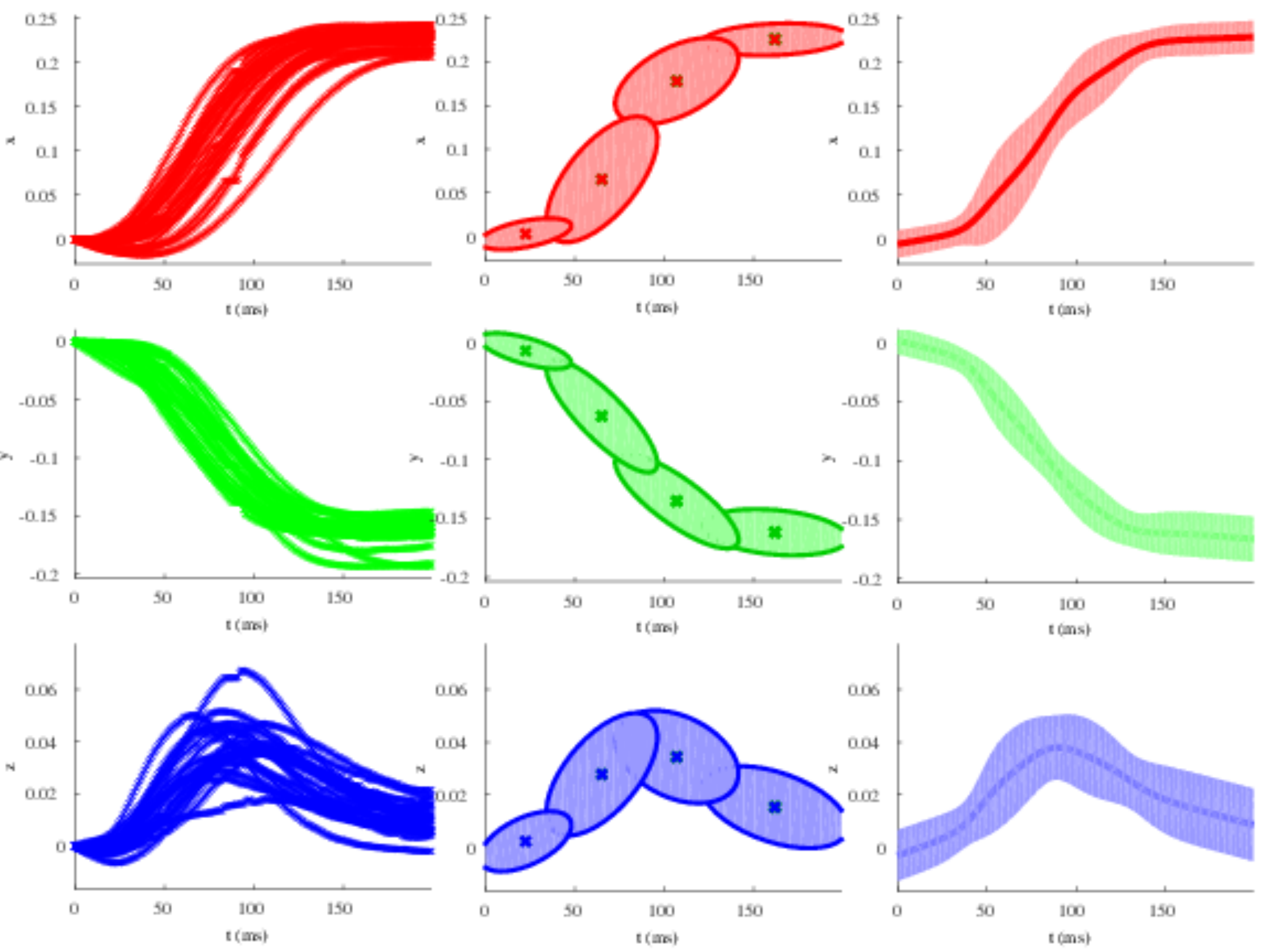}}}
	\caption{Recorded coordinates of human hand performing $P_{R}$ action, representation of corresponding covariance matrices and output from GMR with covariance information.}
	\label{fig:Human_data_and_GMR_A1}
\end{figure}

The left column in Fig.~\ref{fig:Human_data_and_GMR_A1} shows an example of the recorded trajectories of the actor's hand during execution of the $P_{R}$ action. The middle column shows the recorded trajectories encoded in GMM, with covariances matrices represented by ellipses. We use four Gaussian distributions to model the behaviour of the arm trajectory for each Cartesian coordinate. 
This is to take into account the minimum error and the increase of complexity of the problem. Then the signal is reconstructed using Gaussian Mixture Regression (GMR). The new parameters, mean and covariance for each Cartesian coordinate, are defined as in \cite{Calinon07}. The right column represents the GMR output of the signals in bold and the covariance information as the envelope around the bold line.  

\begin{figure}[h]
	\centering
	\framebox{\parbox{3.3in}{\includegraphics[width = 0.48\textwidth]{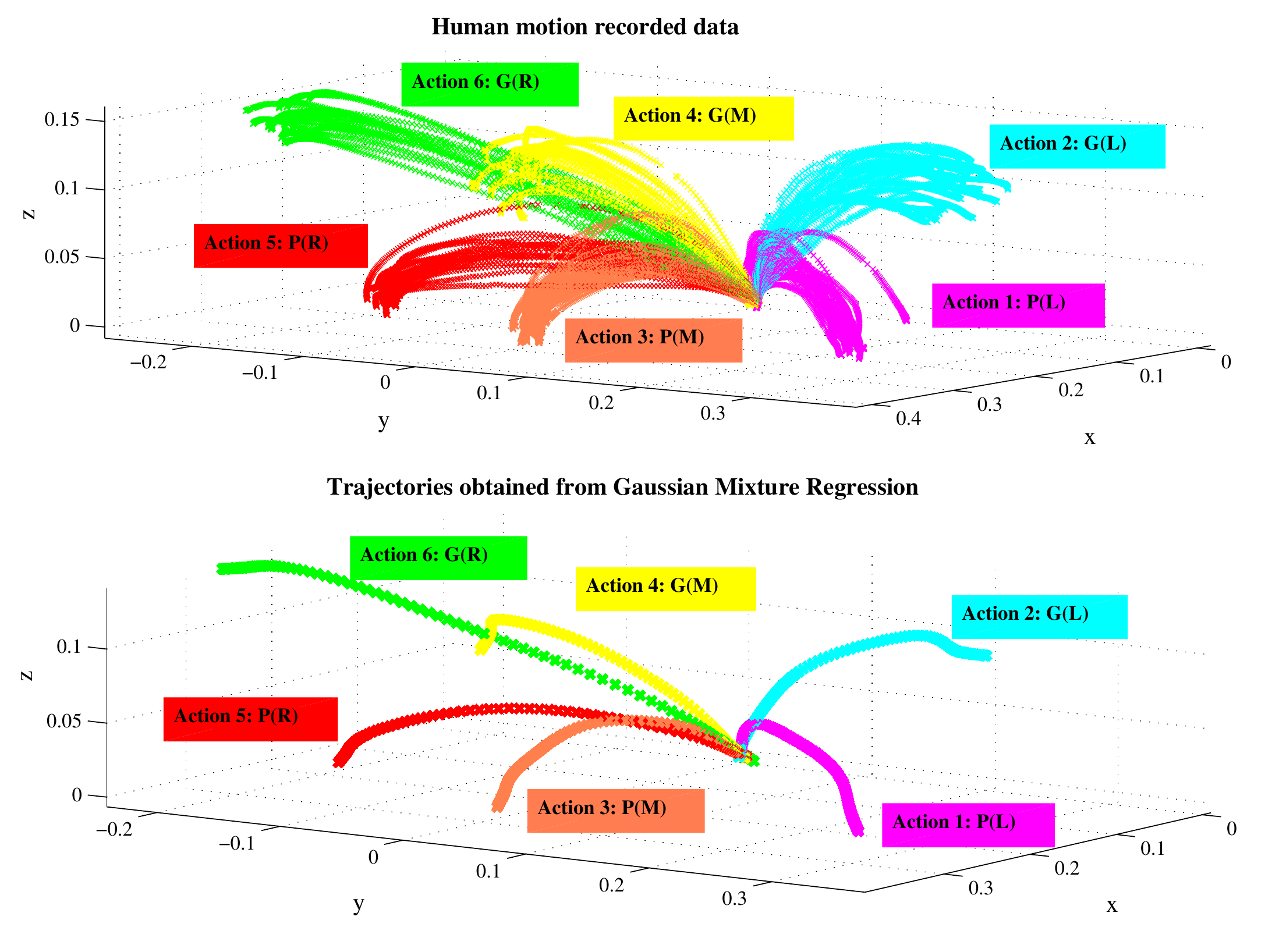}}}
	\caption{Spatial distribution of hand motion for all six actions (top) and corresponding output from GMR (bottom)}
	\label{fig:Human_data_and_GMR}
\end{figure}

The same modelling is done for all the 2 actions and 3 orientations. Fig. \ref{fig:Human_data_and_GMR} - top, shows the spatial distribution of the recorded data for all six actions represented by six different colours. Fig. \ref{fig:Human_data_and_GMR} - bottom, shows the spatial distribution of modelled actions obtained with GMR.

\begin{figure*}[t!]
	\begin{subfigure}{0.195\textwidth}
	\includegraphics[trim={3cm 0 3cm 0},clip, height=0.6\textwidth]{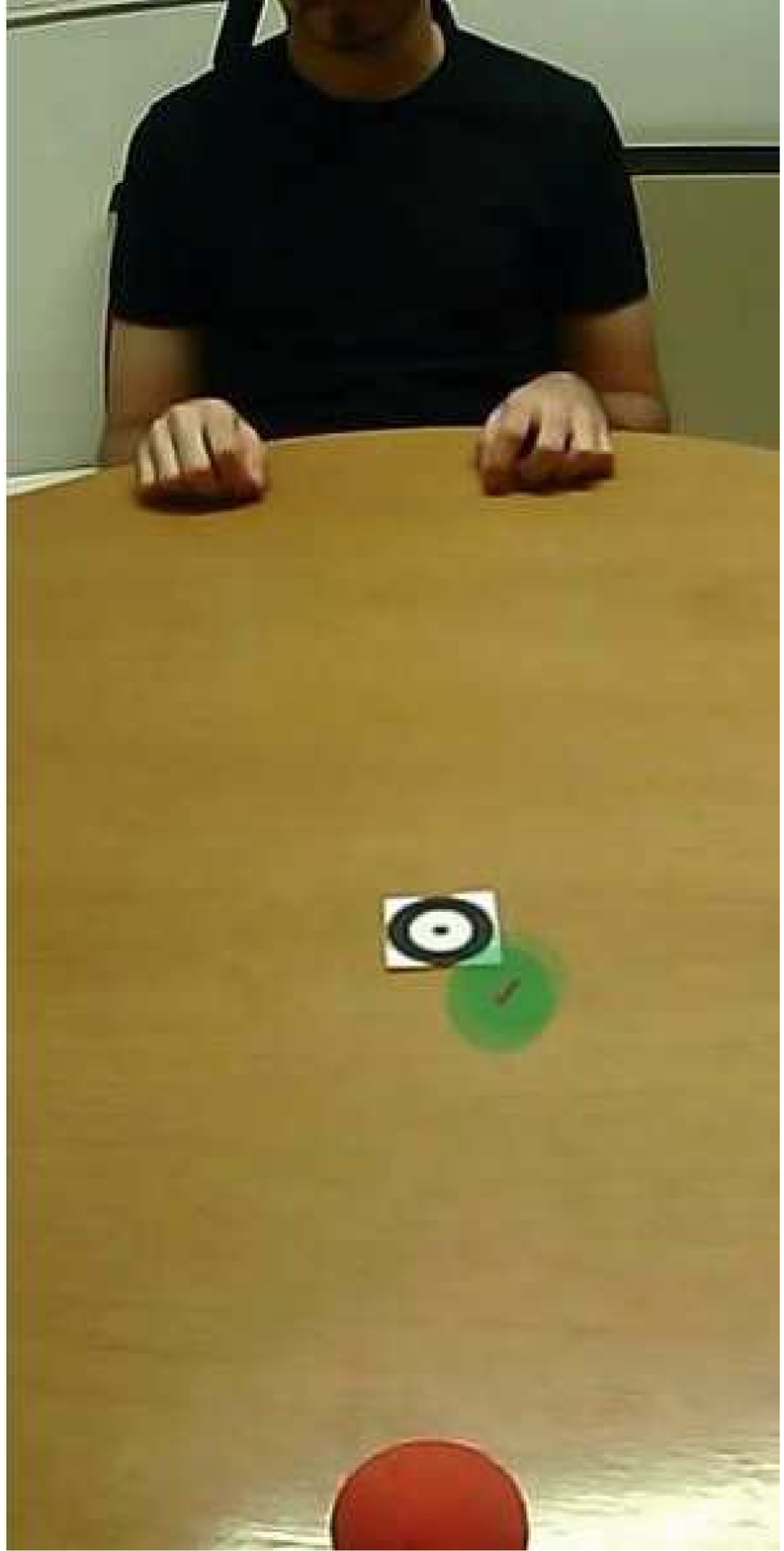}
	\includegraphics[trim={3cm 0 3cm 0},clip, height=0.6\textwidth]{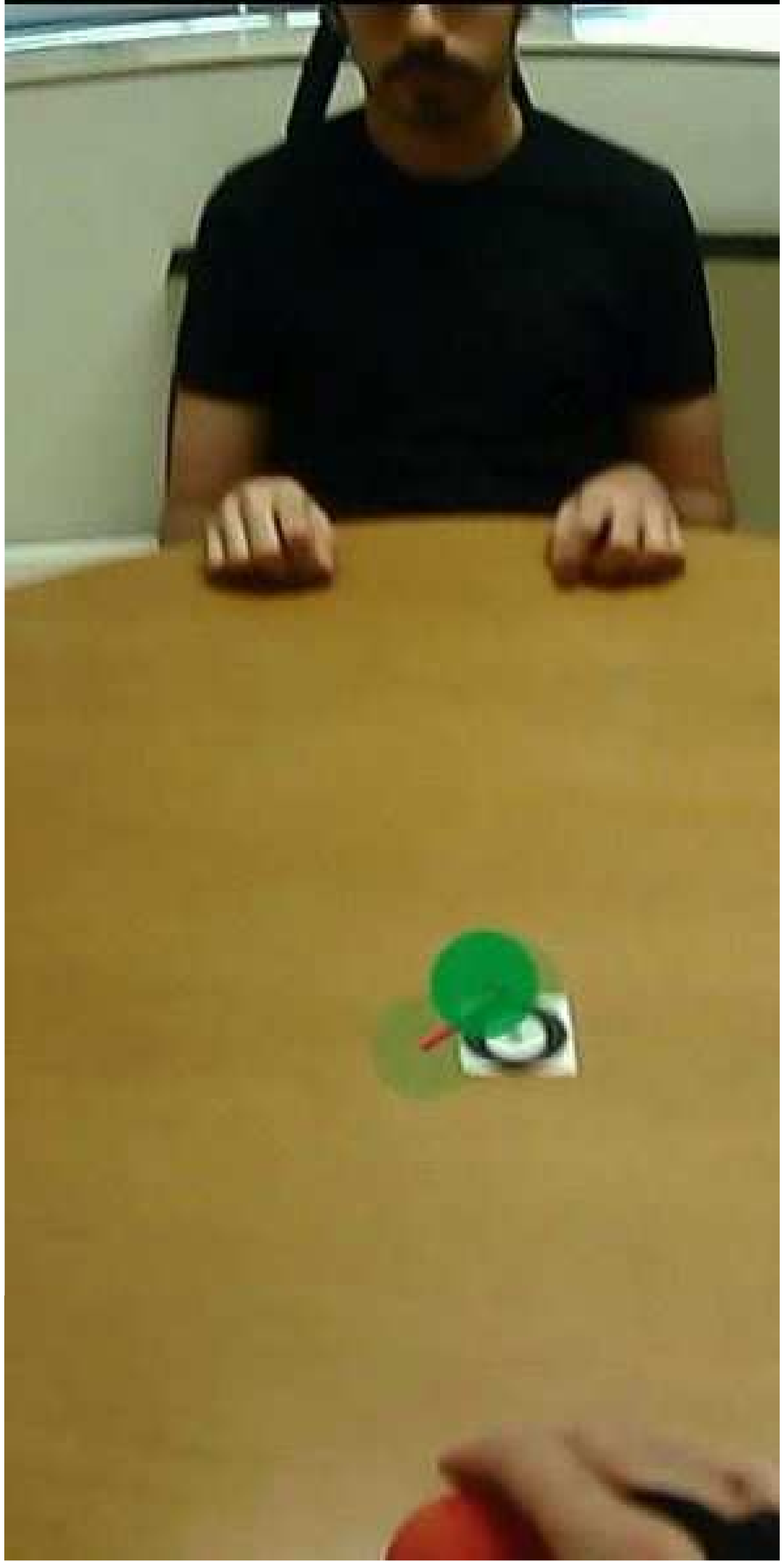}
	\includegraphics[trim={3cm 0 3cm 0},clip, height=0.6\textwidth]{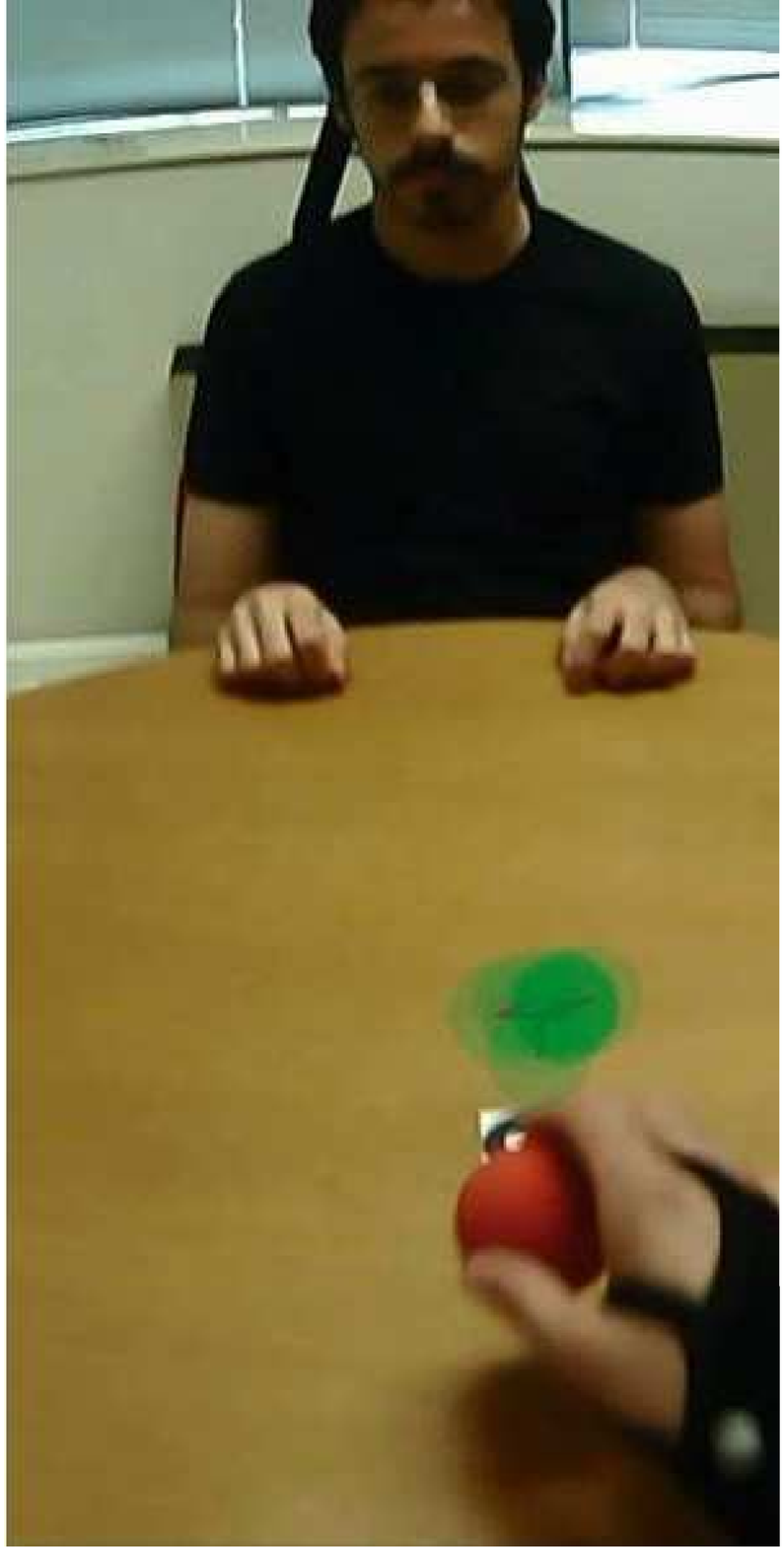}
	\caption{}\label{fig:gaze_seq_a}
	\end{subfigure}
	\begin{subfigure}{0.195\textwidth}
	\includegraphics[trim={3cm 0 3cm 0},clip, height=0.6\textwidth]{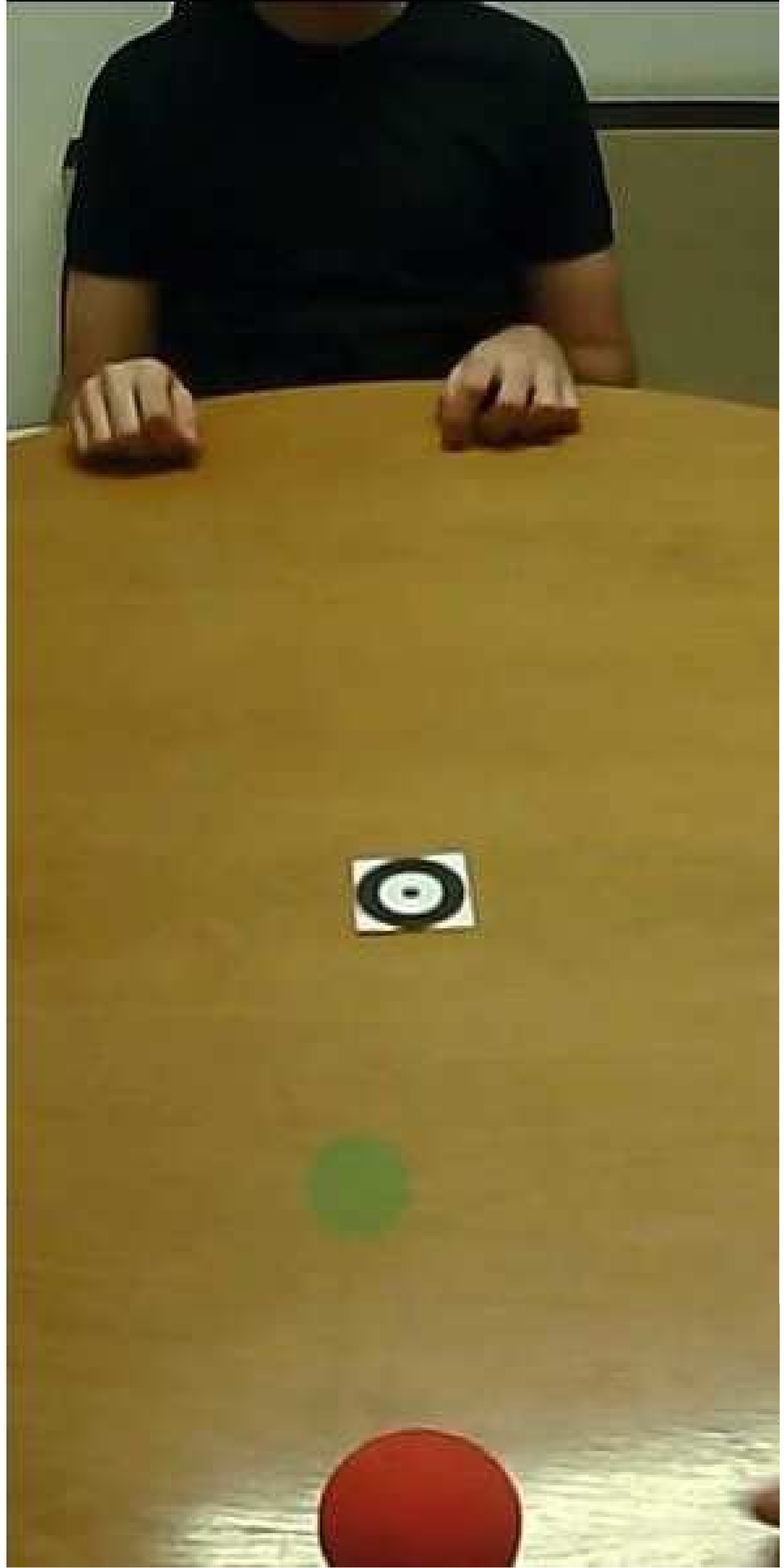}
	\includegraphics[trim={3cm 0 3cm 0},clip, height=0.6\textwidth]{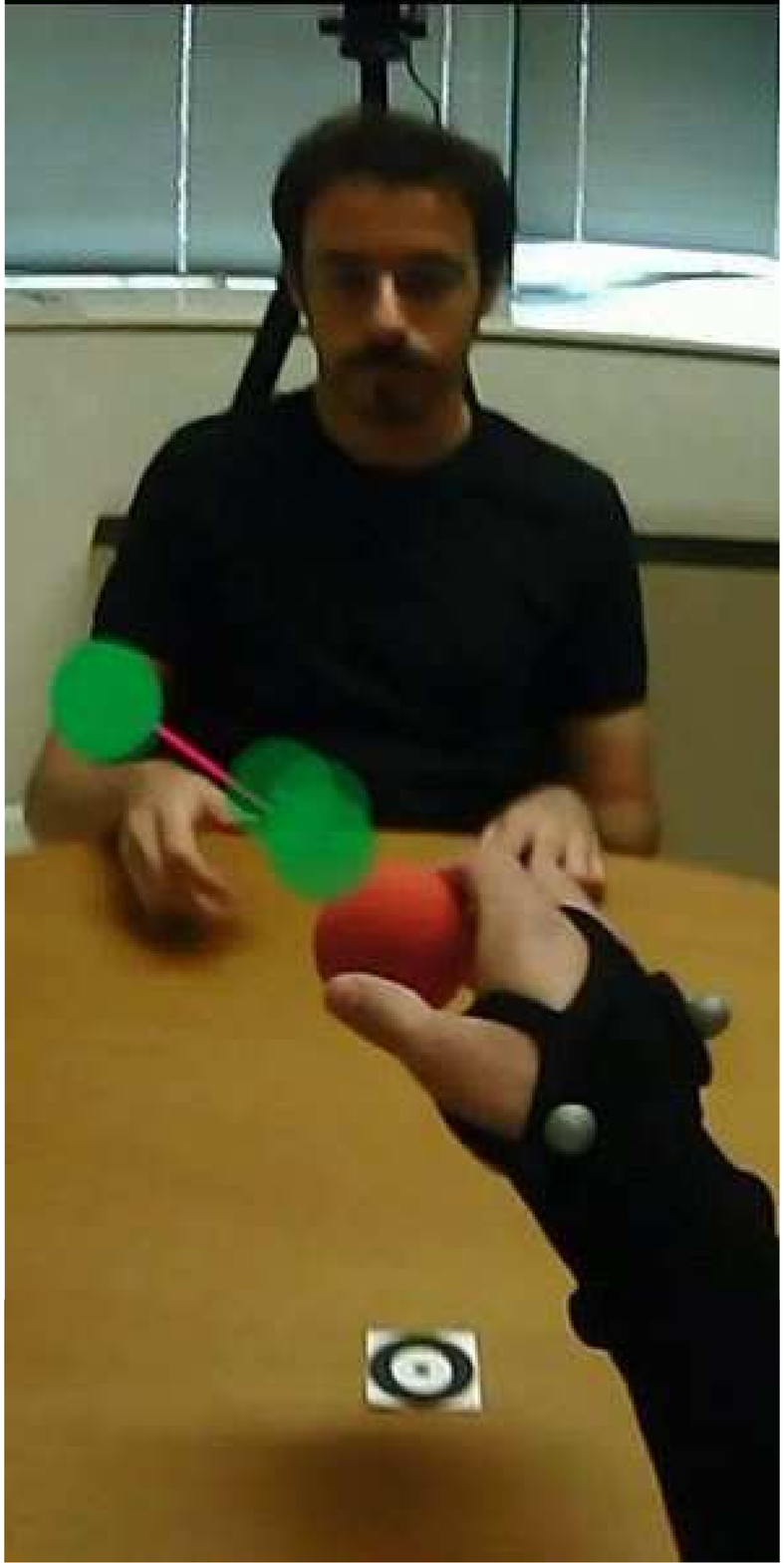}
	\includegraphics[trim={3cm 0 3cm 0},clip, height=0.6\textwidth]{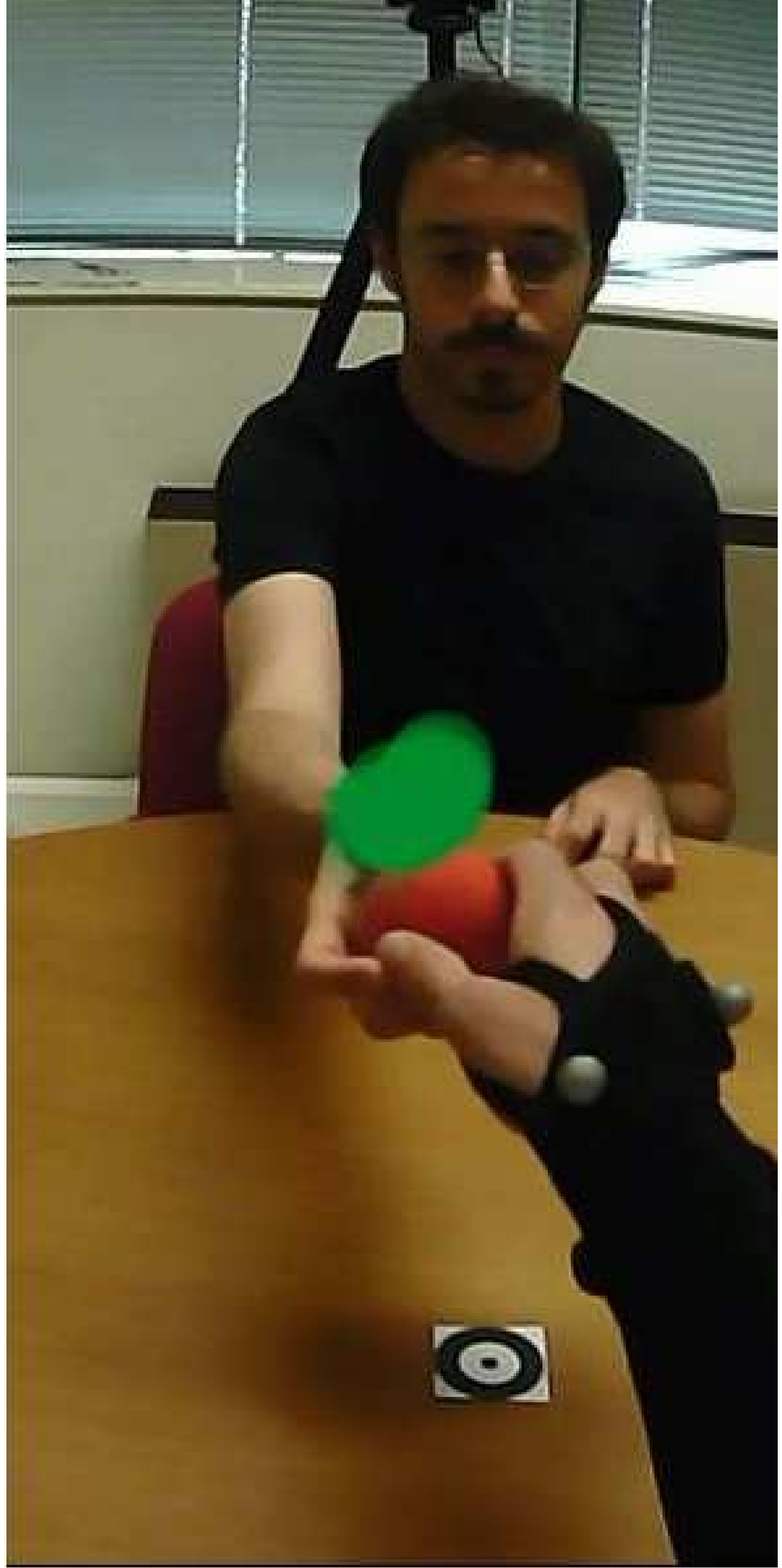}
	\caption{}\label{fig:gaze_seq_b}	
	\end{subfigure}
	\begin{subfigure}{0.195\textwidth}
	\includegraphics[trim={3cm 0 3cm 0},clip, height=0.6\textwidth]{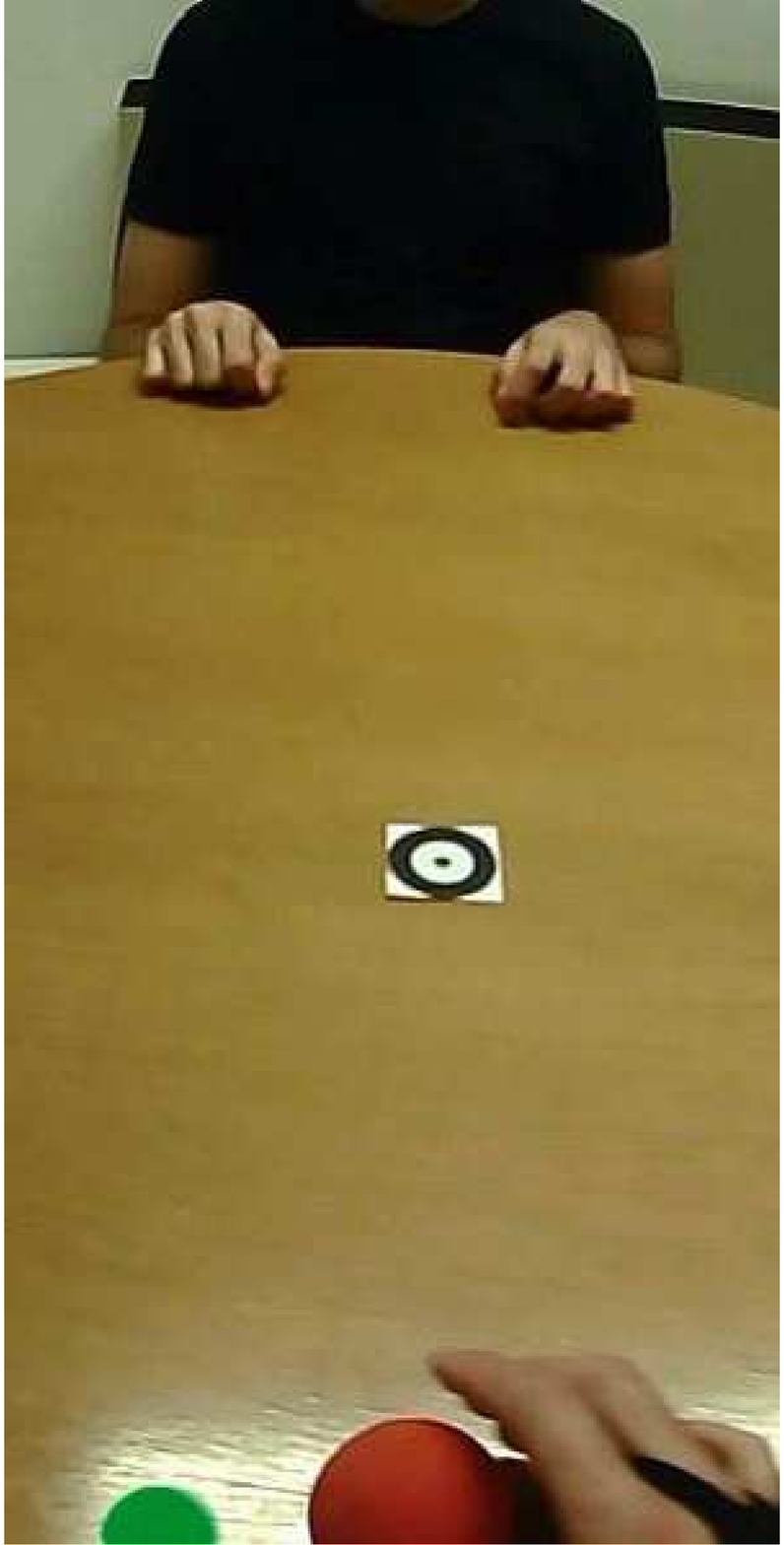}
	\includegraphics[trim={3cm 0 3cm 0},clip, height=0.6\textwidth]{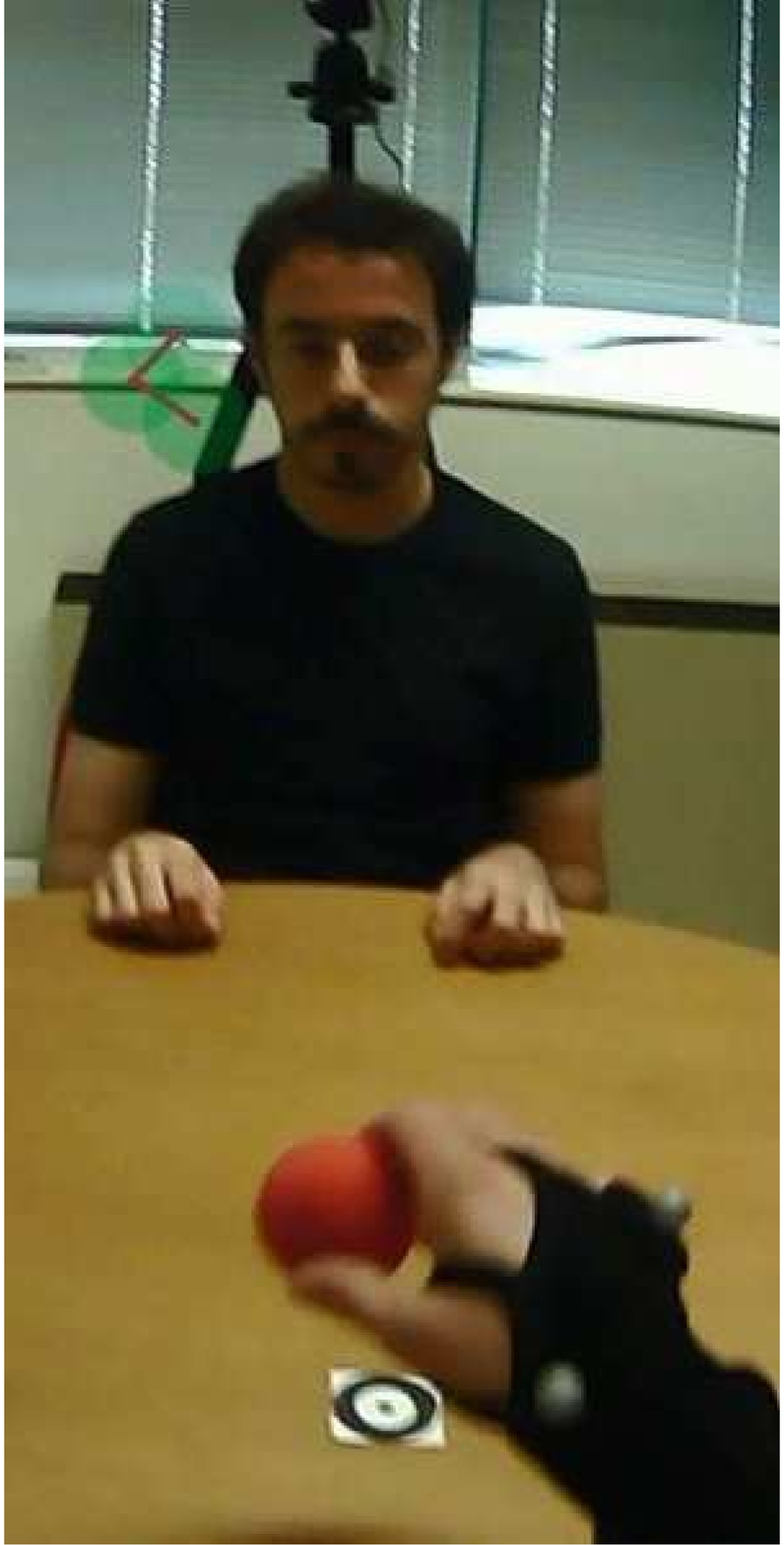}
	\includegraphics[trim={3cm 0 3cm 0},clip, height=0.6\textwidth]{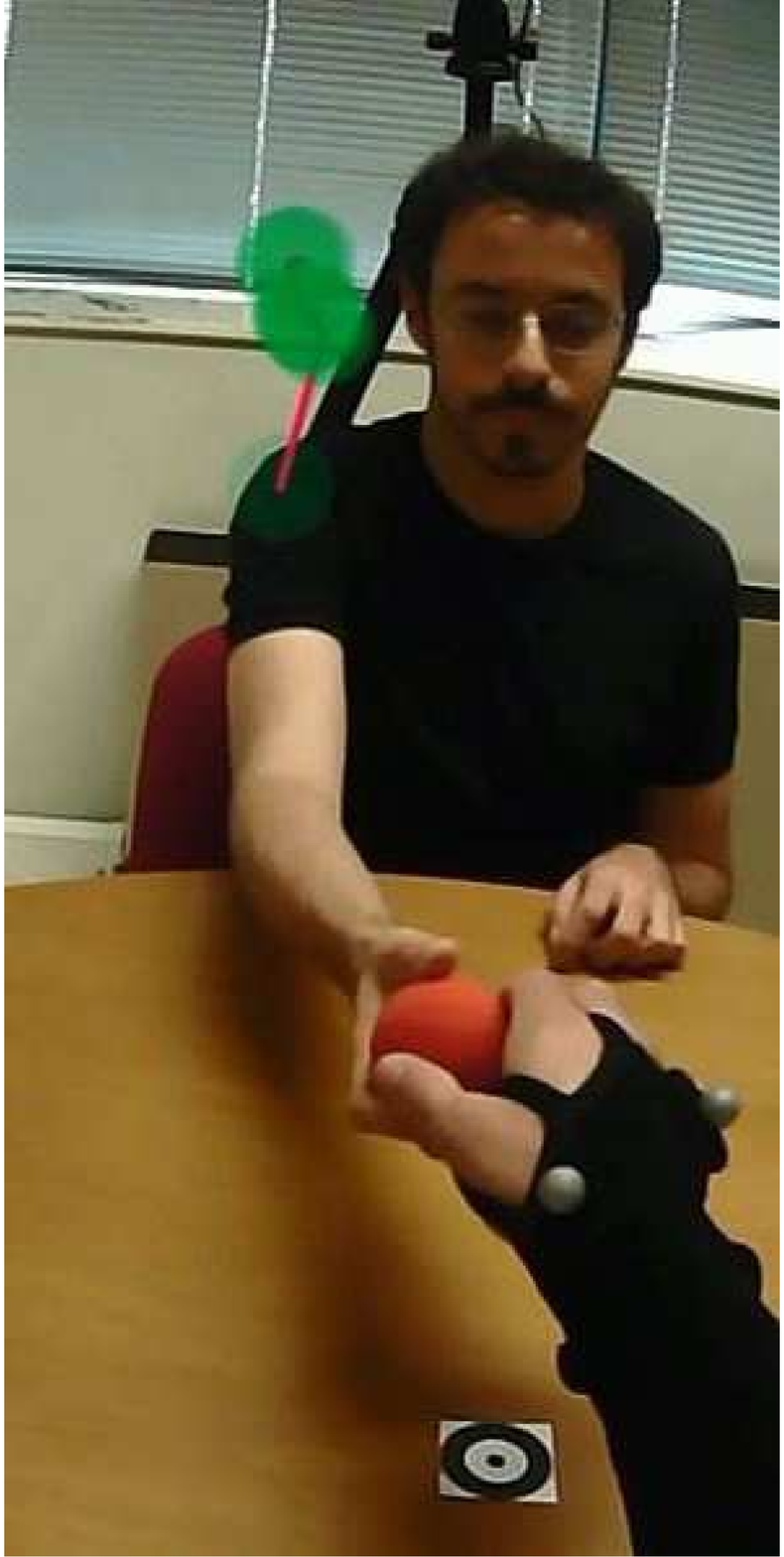}
	\caption{}\label{fig:gaze_seq_c}
	\end{subfigure}
	\begin{subfigure}{0.195\textwidth}
	\includegraphics[trim={3cm 0 3cm 0},clip, height=0.6\textwidth]{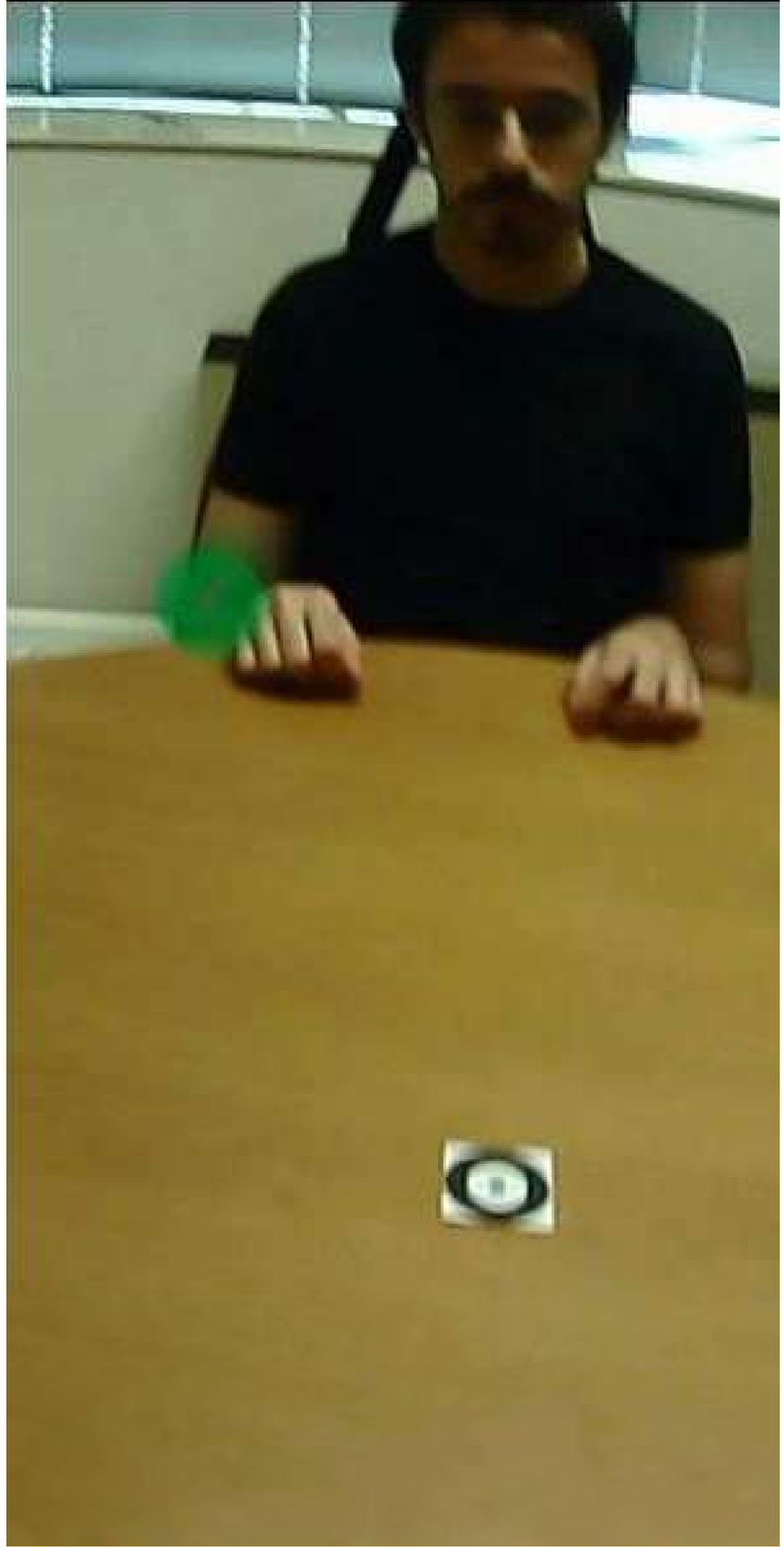}
	\includegraphics[trim={3cm 0 3cm 0},clip, height=0.6\textwidth]{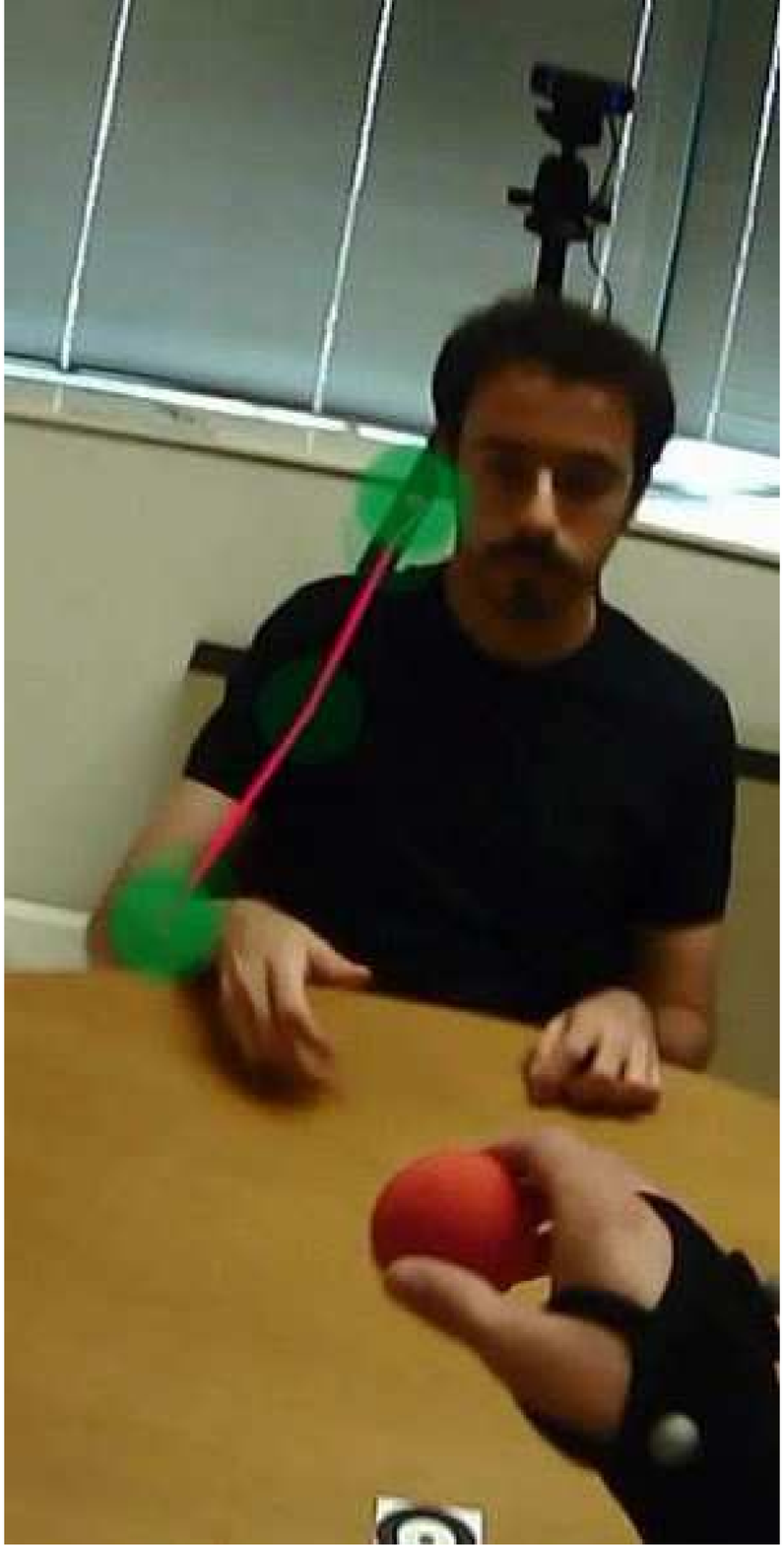}
	\includegraphics[trim={3cm 0 3cm 0},clip, height=0.6\textwidth]{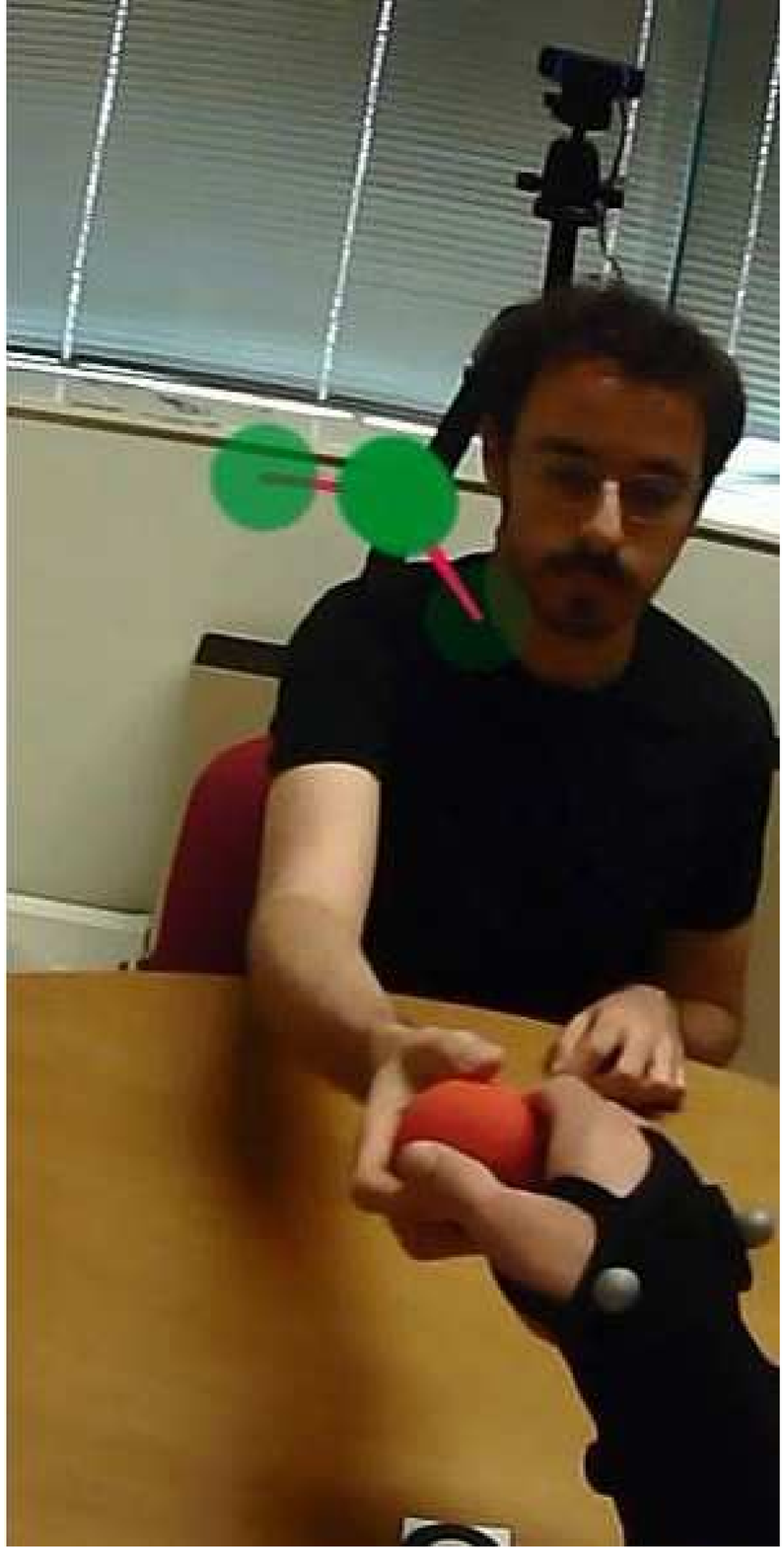}
	\caption{}\label{fig:gaze_seq_d}
	\end{subfigure}
	\begin{subfigure}{0.195\textwidth}
	\includegraphics[trim={3cm 0 3cm 0},clip, height=0.6\textwidth]{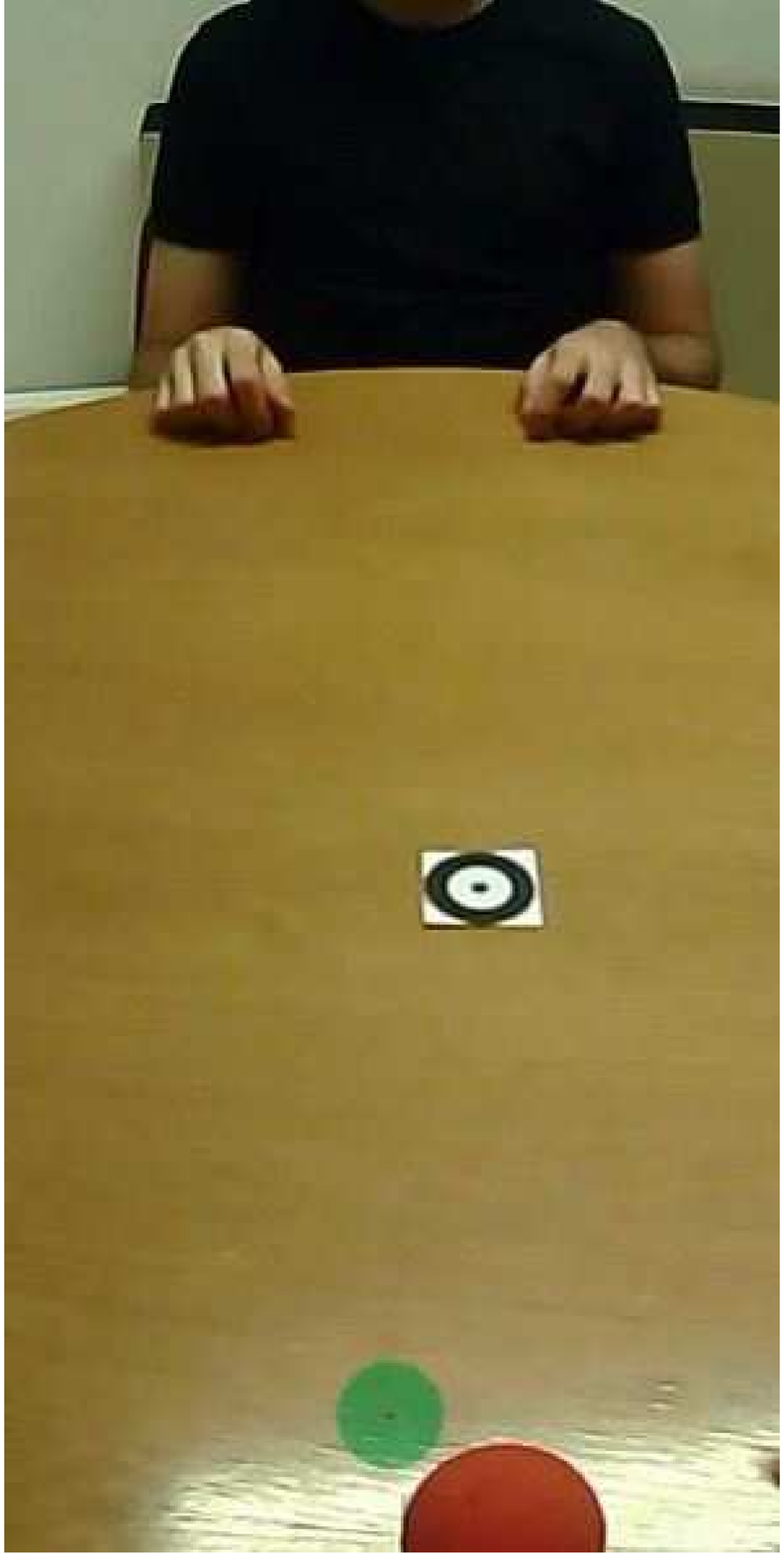}
	\includegraphics[trim={3cm 0 3cm 0},clip, height=0.6\textwidth]{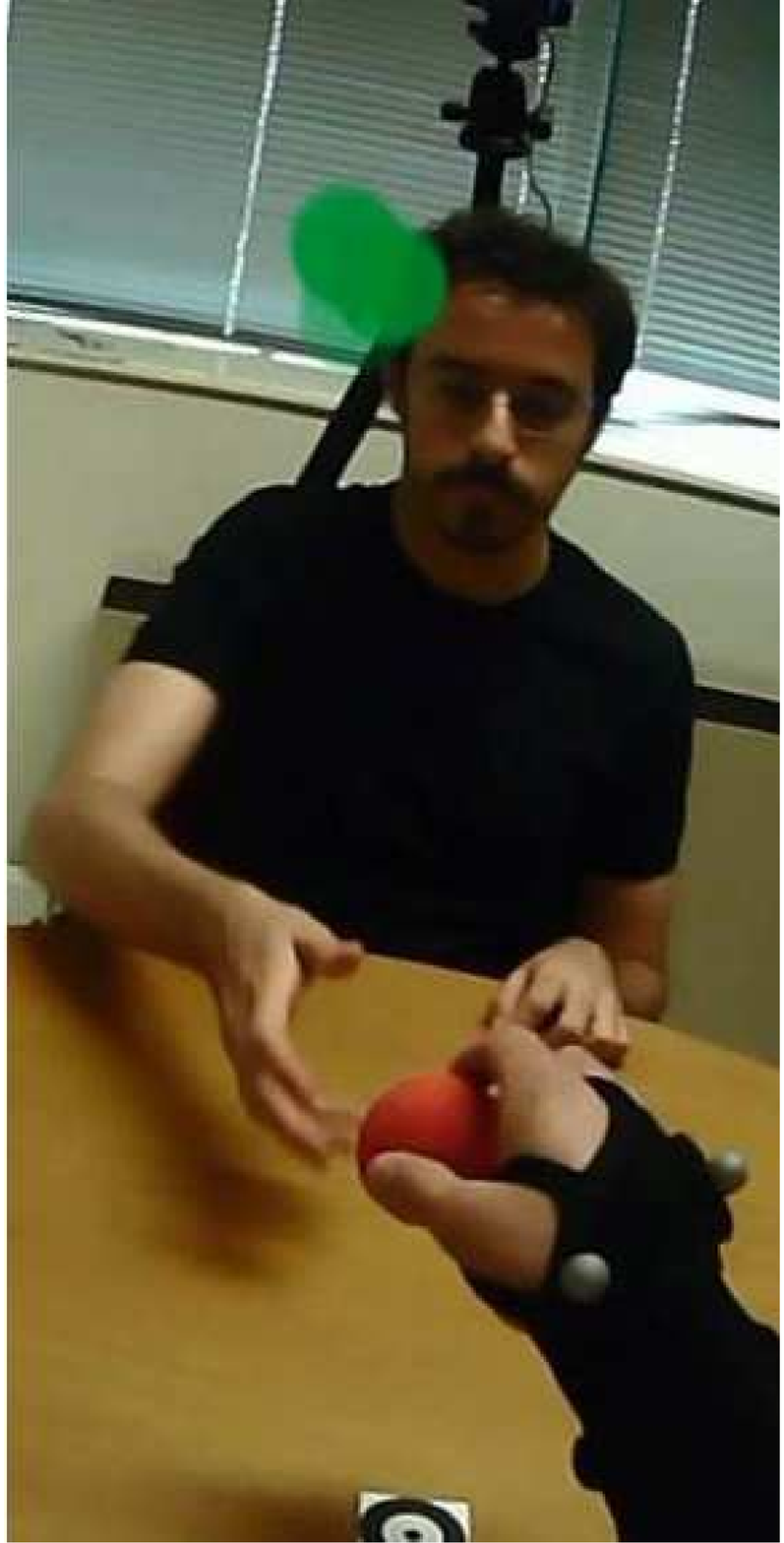}
	\includegraphics[trim={3cm 0 3cm 0},clip, height=0.6\textwidth]{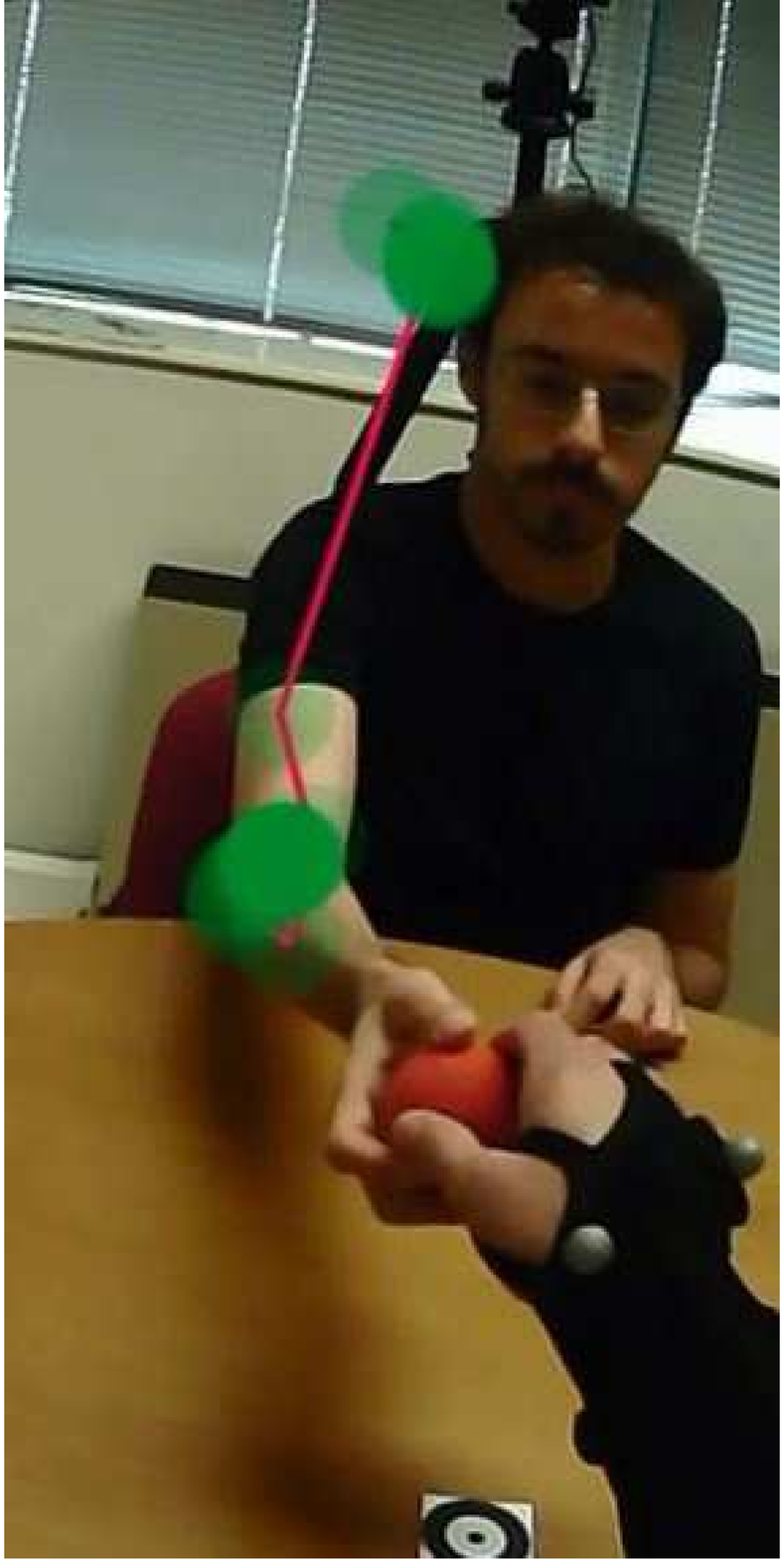}
	\caption{}\label{fig:gaze_seq_e}
	\end{subfigure}
	\caption{Sequence of images of spatiotemporal distribution of fixation point for \textit{placing} and \textit{giving} actions. Subgroup (a) is related to action $P_{M}$. The actor only fixates the center marker which is the end-goal point for the action. Subgroups (b)-(e) correspond to action $G_{M}$. The actor changes fixation point in 4 different patterns: (b) actor's only fixates the hand of the subject in front; (c)  only fixating the subject in front; (d) it begins by fixating the subject's hand and it ends by fixating the subject's eyes; (e) it fixates the subject's eyes in the beginning and it ends the fixation by looking at the subject's hand. }\label{fig:gaze_seq}
\end{figure*}
\begin{figure*}[htpb]
	\includegraphics[trim={2cm 0 3cm 0},clip, width=0.19\textwidth]{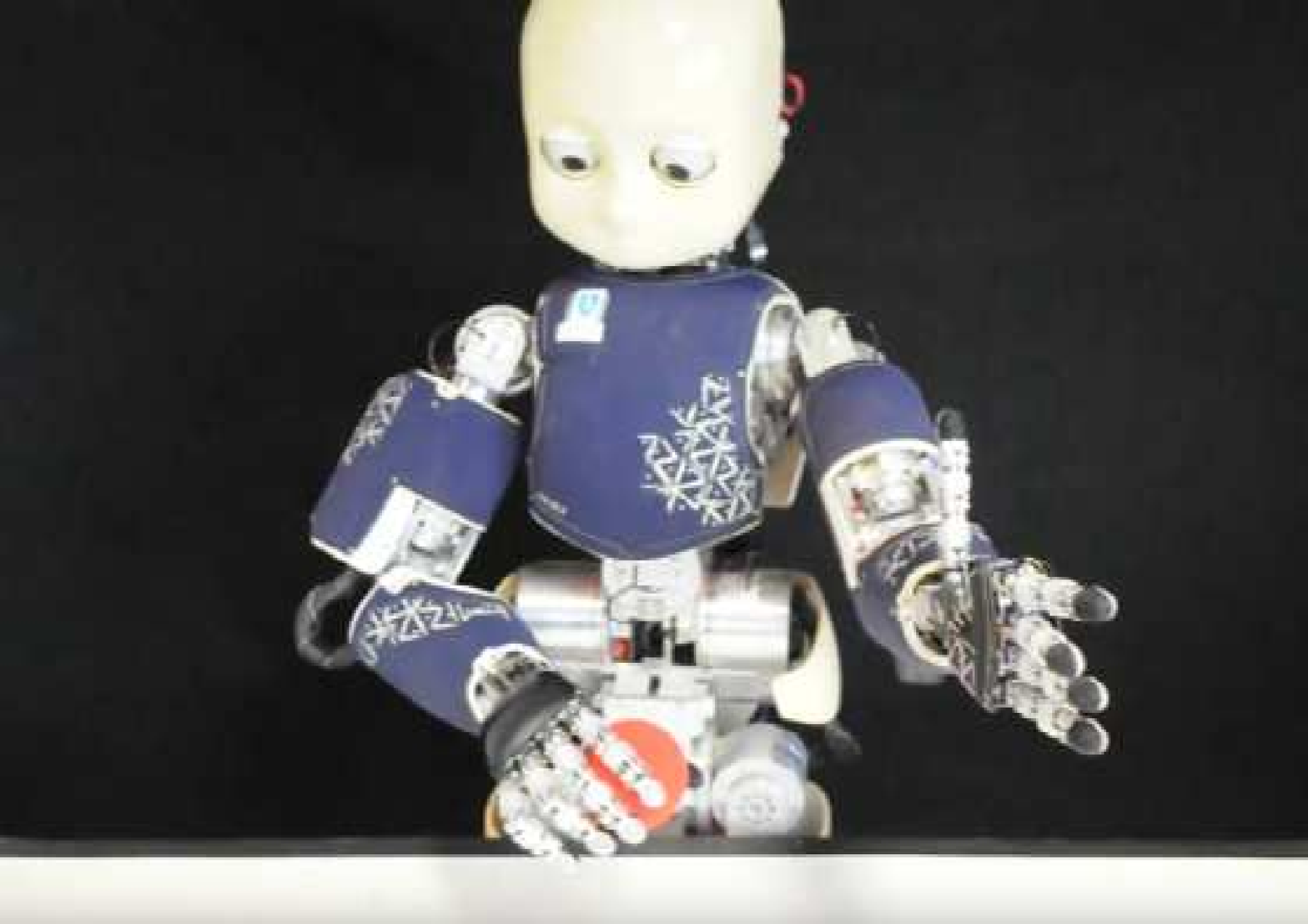} \hfill
	\includegraphics[trim={2cm 0 3cm 0},clip, width=0.19\textwidth]{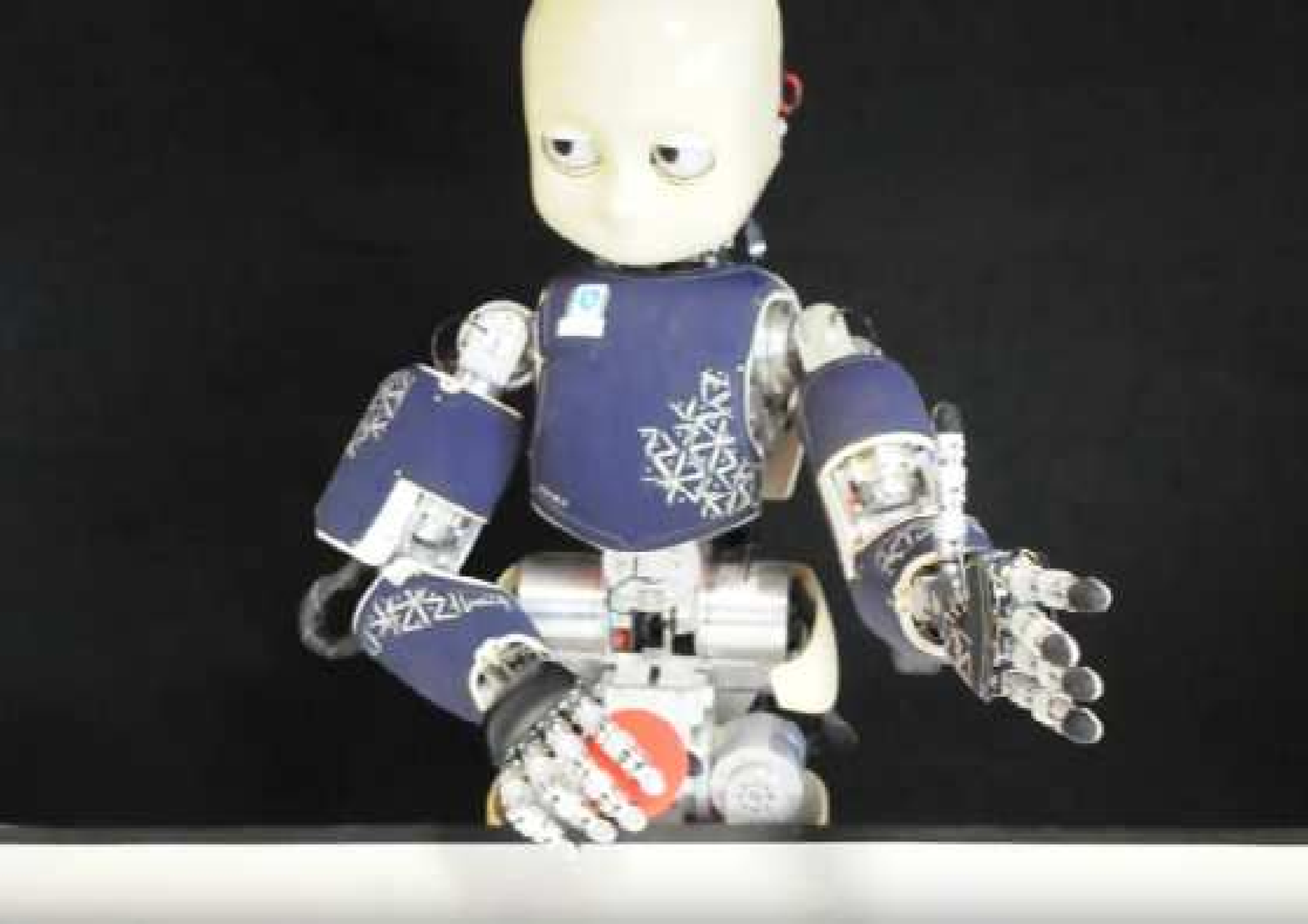} \hfill
	\includegraphics[trim={2cm 0 3cm 0},clip, width=0.19\textwidth]{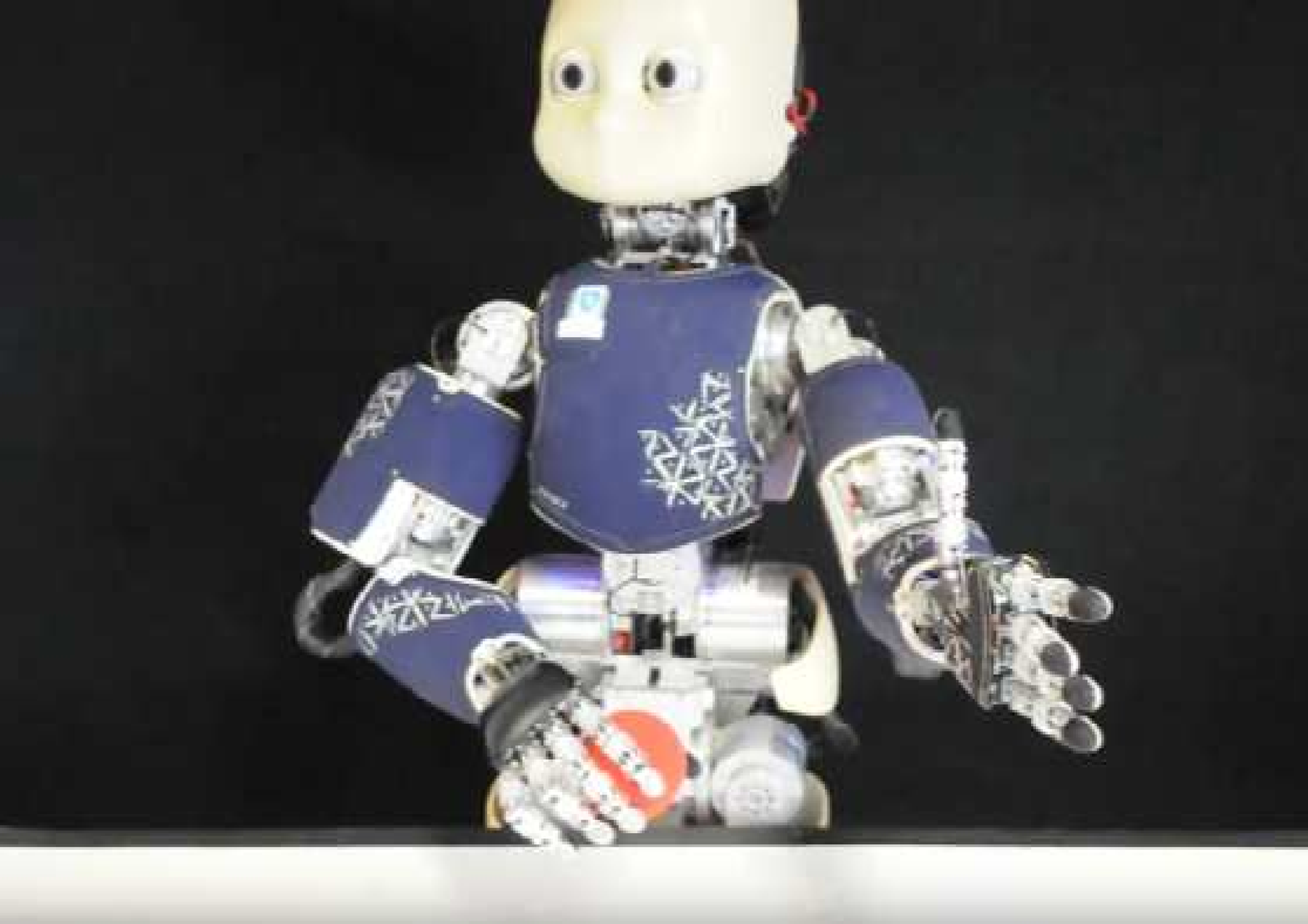} \hfill
	\includegraphics[trim={2cm 0 3cm 0},clip, width=0.19\textwidth]{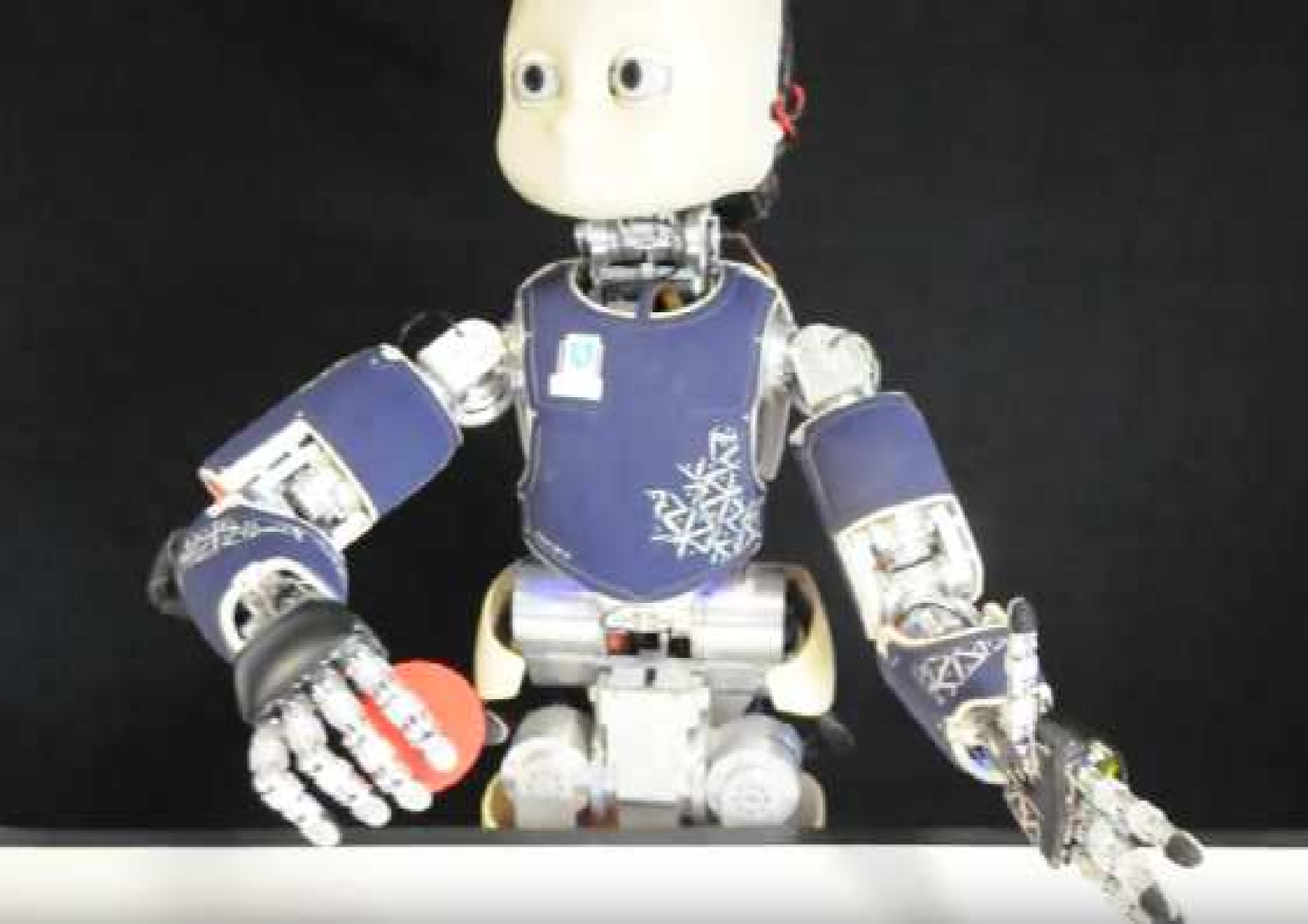} \hfill
	\includegraphics[trim={2cm 0 3cm 0},clip, width=0.19\textwidth]{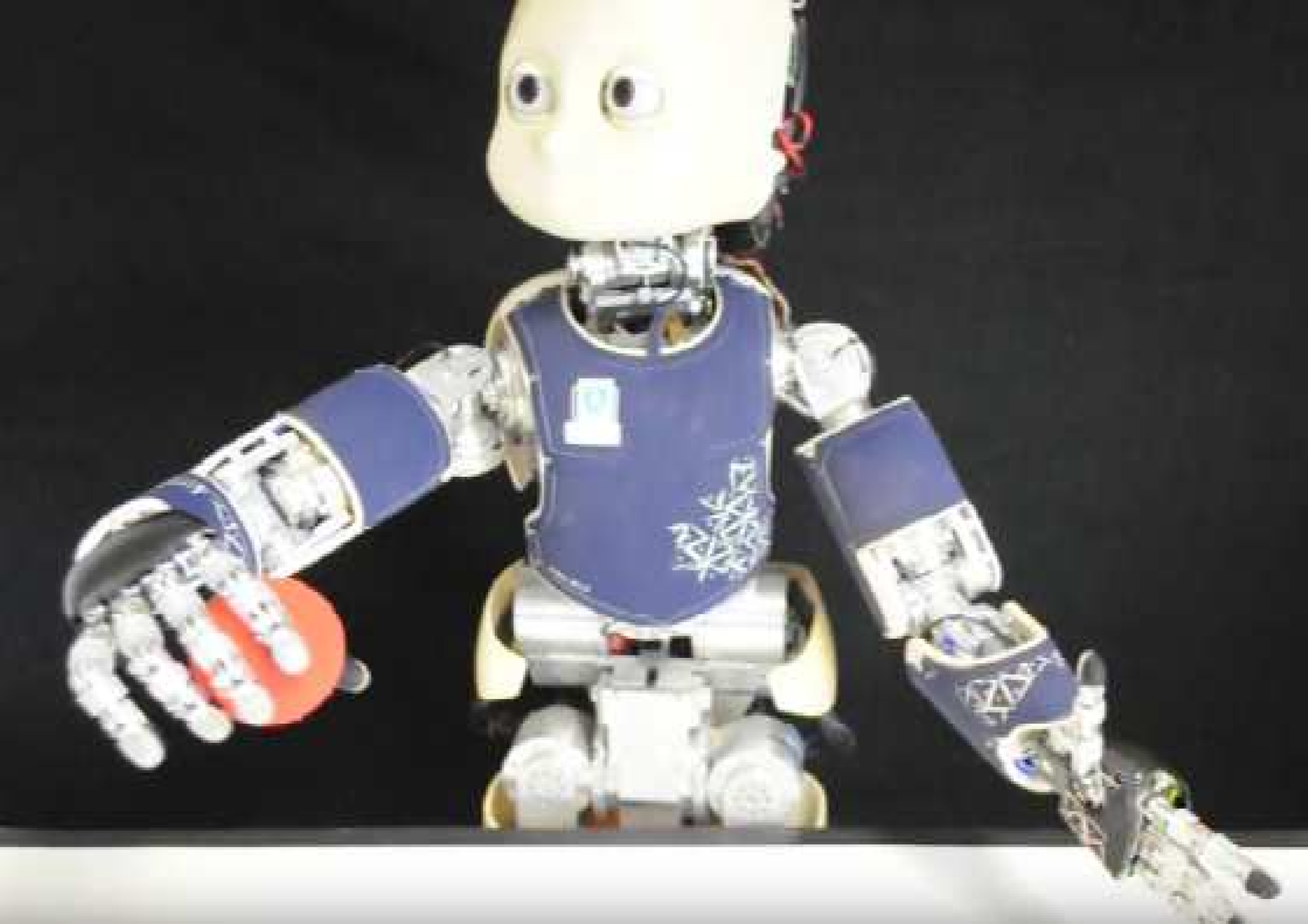} 
	
	\hspace{0.5cm}
	
	\includegraphics[clip, width=0.19\textwidth]{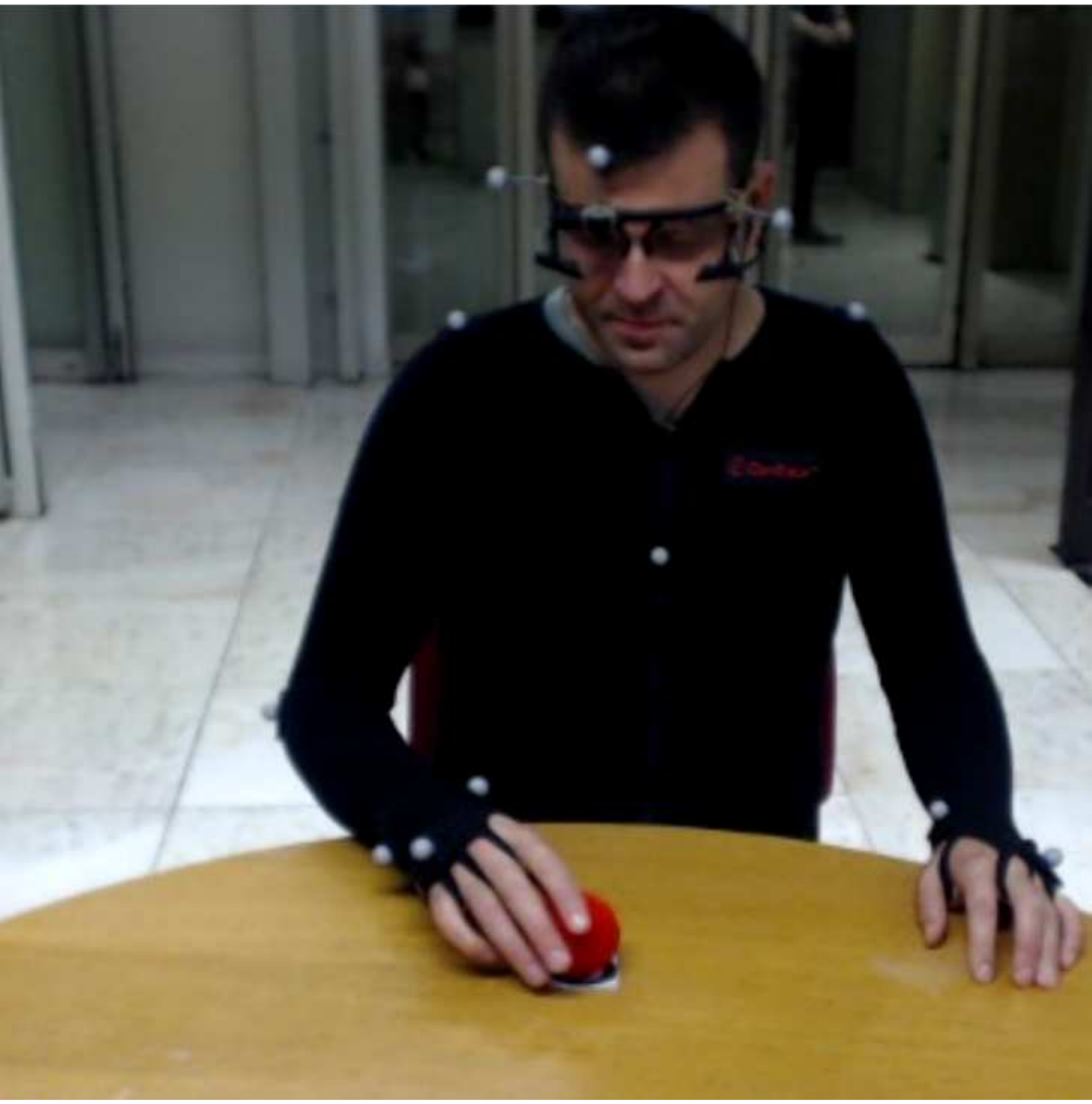}\hfill
	\includegraphics[clip, width=0.19\textwidth]{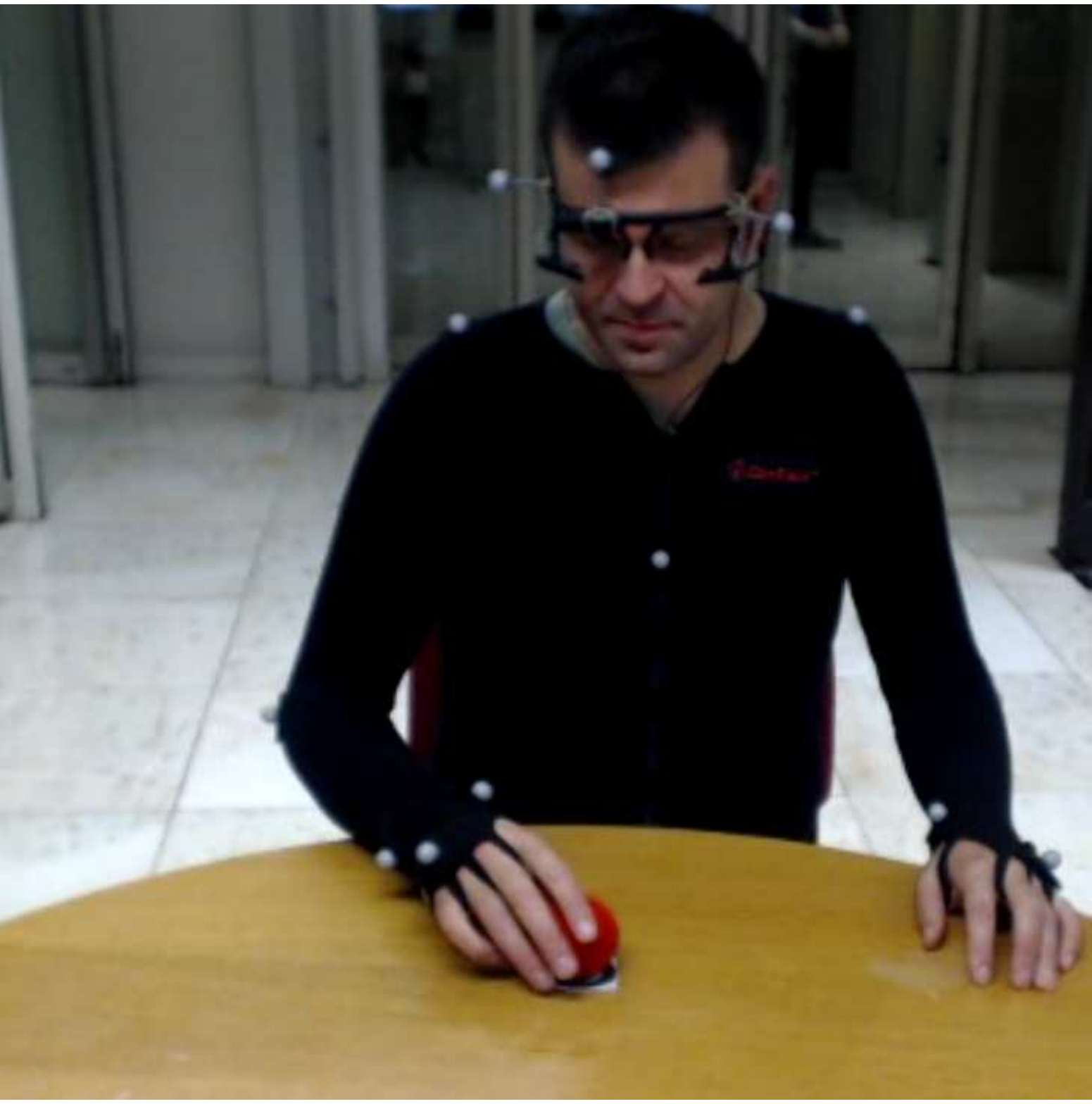}\hfill
	\includegraphics[clip, width=0.19\textwidth]{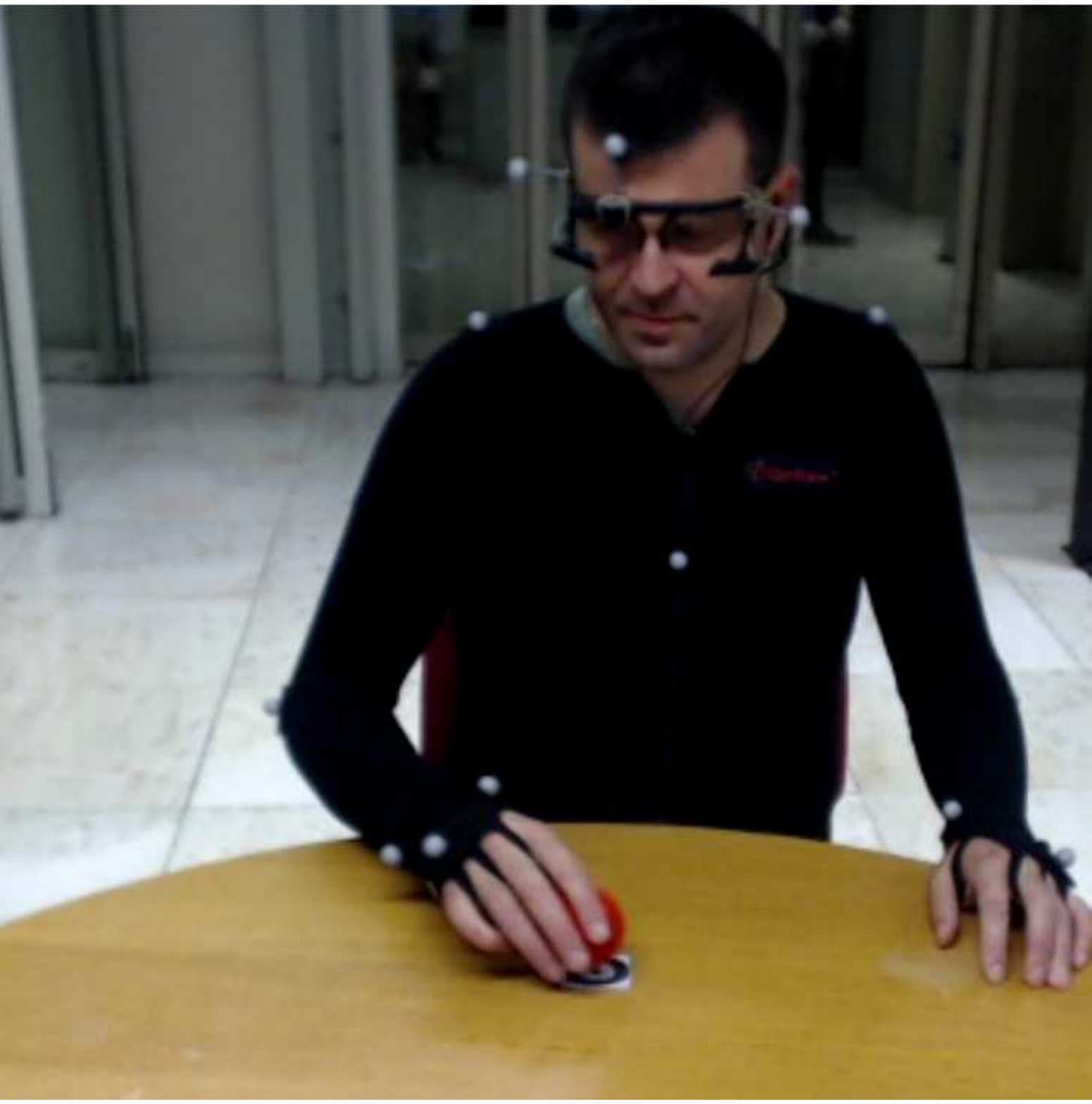}\hfill
	\includegraphics[clip, width=0.19\textwidth]{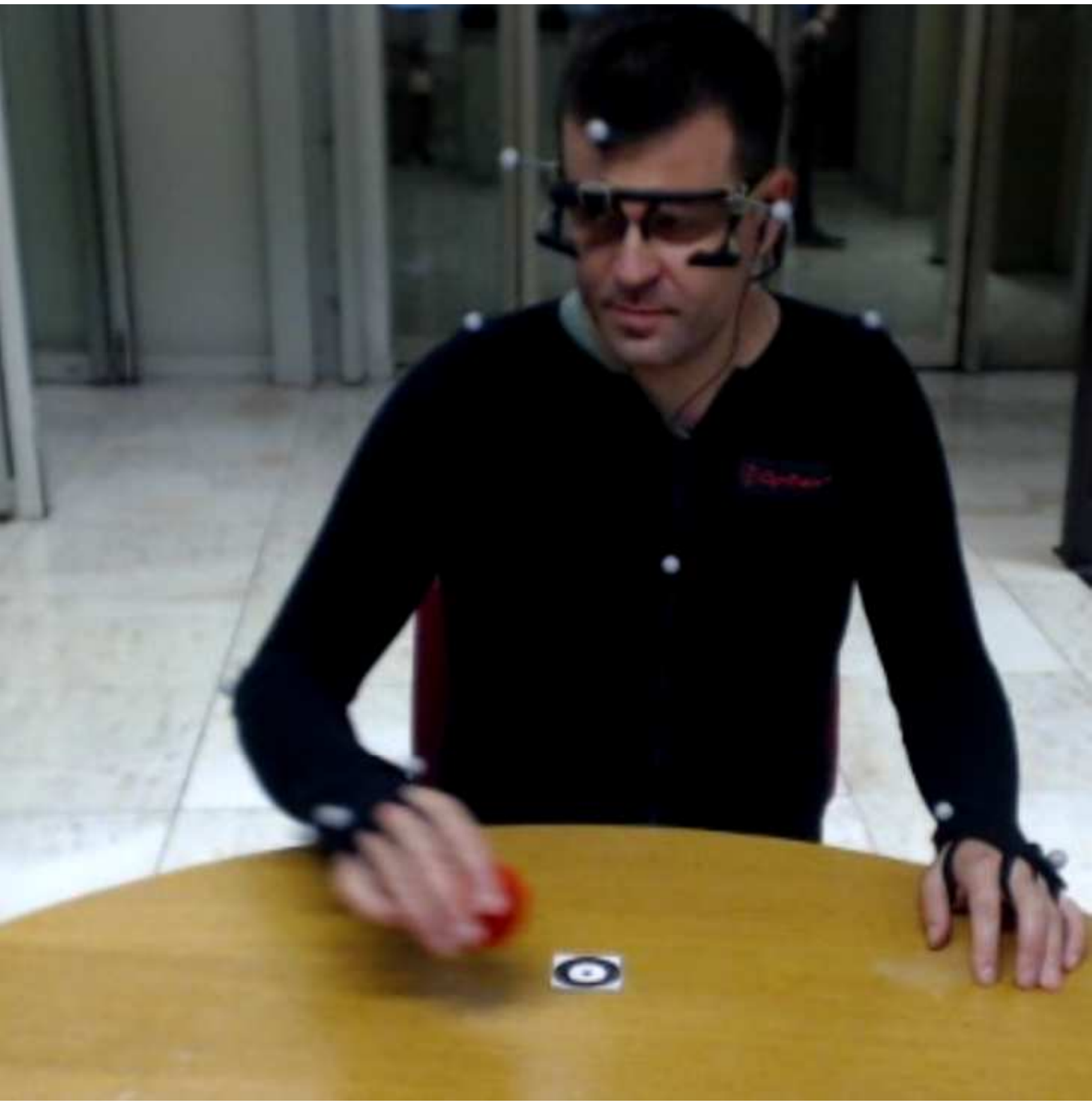}\hfill
	\includegraphics[clip, width=0.19\textwidth]{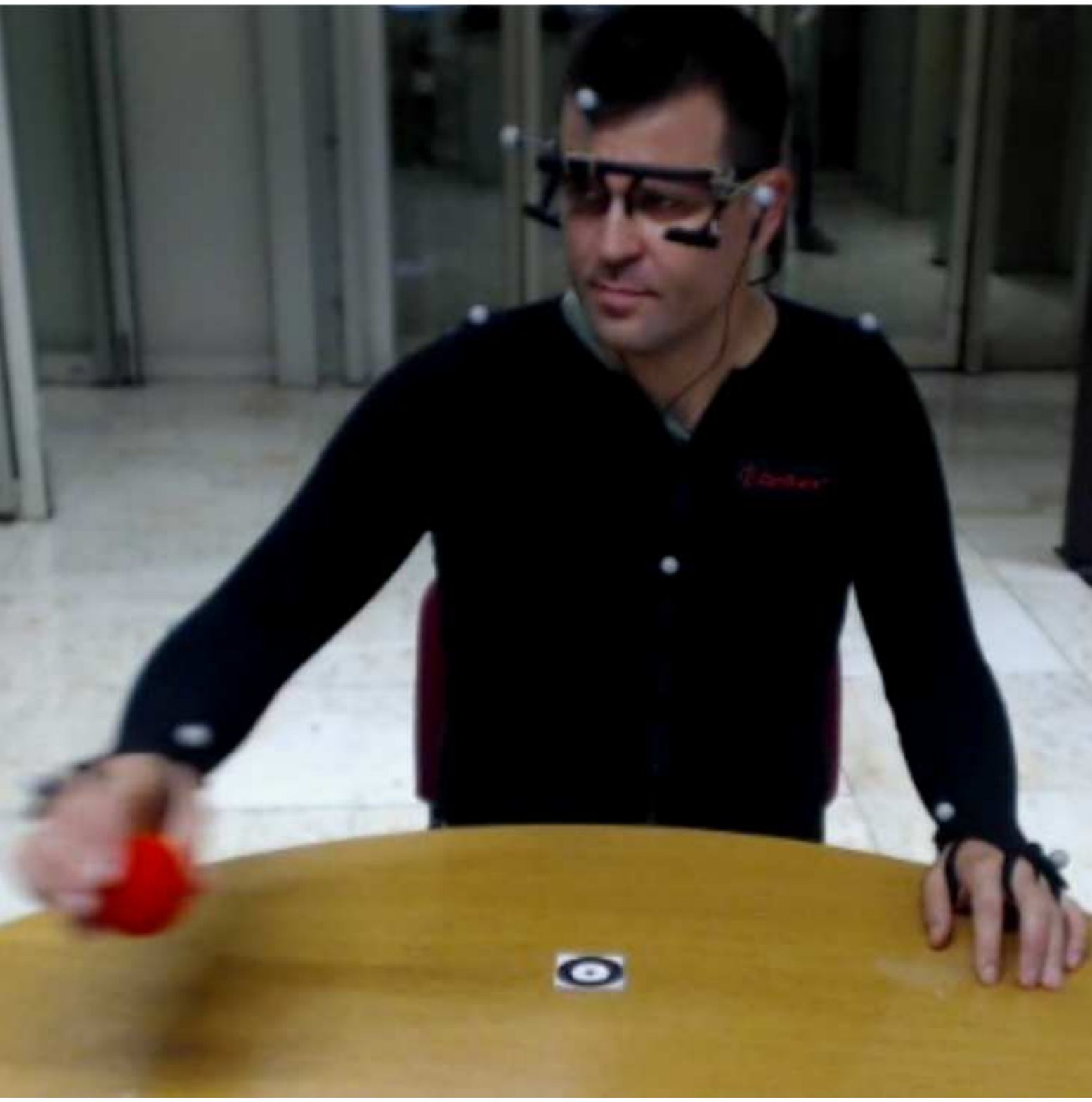}\hfill
\caption{The sequence of images of a robot (top) and an actor (bottom) performing the  $G_{R}$ action. The first sequence is the initial point for both the actor and the robot. The second stage corresponds to when the short video stops at the video fraction 'G'. The third is at video fraction 'G+H'. Forth and fifth sequences are for the final two video fractions, corresponding to the arm motion. }\label{fig:rob_seq}
\end{figure*}

Moon et al.\cite{moon2014meet} observed that the human eye gaze exhibits a switching  behaviour during \textit{giving} actions. This was observed in a HHI experiment scenario where two humans are \textit{giving} a bottle to each other. 
The work has several shortcomings. First, the experiment can not guarantee that gaze behaviour occurs in general settings. Once the human knows which action will take place, there is no need to infer the action from non-verbal communication.  Secondly, the analysis of the different gaze behaviours was done empirically (manually labelling the videos). In our dataset, we have measured fixation points, the actual points of interest in a handover task, and the duration of eye gaze between each switching behaviour. As future work, we use this information to design a detailed biologically-inspired, eye-gaze controller for HRI scenarios.

In the HRI experiment of \cite{moon2014meet}, when the robot gaze fixations switches from the human's face to the handover position, it does not improve the speed of the human reaching time, but it does improve the perception of the interaction. This corroborates the findings in \cite{zheng2015impacts}. Our work studies the same behaviour, using a humanoid robot and eye-gaze cues extracted from the HHI experiment.

The information collected to model the human gaze behaviour, was acquired with an eye-tracking system (Pupil-Labs). Fig. \ref{fig:gaze_seq} shows five different cases of the spatio-temporal distribution of the fixation point marked with a green circle. Fig. \ref{fig:gaze_seq_a} shows the spatio-temporal distribution of fixation points for the $P_{M}$ \textit{placing} action in which the green circle is concentrated around the goal position of the red ball. 

Fig. \ref{fig:gaze_seq_b}-\ref{fig:gaze_seq_e} show the spatio-temporal distributions of the fixation points during $G_{M}$ \textit{giving} action when the actor was fixating: (i) only the hand of the person, (ii) only the face of the person, (iii) first the hand and then the face, and (iv) first the face and then the hand. From this observed behaviour, we designed a controller that will generate an equivalent switching behaviour of the fixation point, i.e. a qualitatively similar eye-gaze behaviour.

The robot gaze controller was implemented as a state-machine that (qualitatively) replicates the gaze shift behaviour observed during human-human interaction. The controller's initial state is the starting location of the ball. Then, depending on the action, there is a state transition to the final location of the ball (\textit{placing}) or a switch between two states: (i) face of the person, (ii) handover location, (\textit{giving}). The desired fixation point is input to the coupled eye-head controller that executes saccadic eye movements, followed by the coordinated motion of the eye/neck joints. Fig. \ref{fig:rob_seq} shows the sequence of images, during the execution of the $G_{R}$ action by the iCub robot and the corresponding images of the actor, when the actor looks first to hand of the other person and then switches to the face.

The validation of the controller is presented in Section \ref{sec:robots_cues}. The reference arm trajectory is generated with a GMR and the arm's joints are controlled with a minimum jerk Cartesian controller. The robot eye controller was based on the qualitative analysis of the human gaze behaviour and the eye's and neck joints are simultaneously controlled using Cartesian 6-DOF gaze controller \cite{roncone2016cartesian}.

\section{Reading the Intentions of Robots}\label{sec:robots_cues}

To study the readability of robot's intention, we prepared a second questionnaire with the same set of actions performed by a robot. To assess the relative importance of the different non-verbal (eye, head, arm) cues we have added new conditions: (i) blurring the eyes in the video, and (ii) blurring the entire head.

\begin{figure}[t]
	\centering
	\begin{subfigure}{0.49\textwidth}
		\includegraphics[width = 0.99\textwidth, height=5cm]{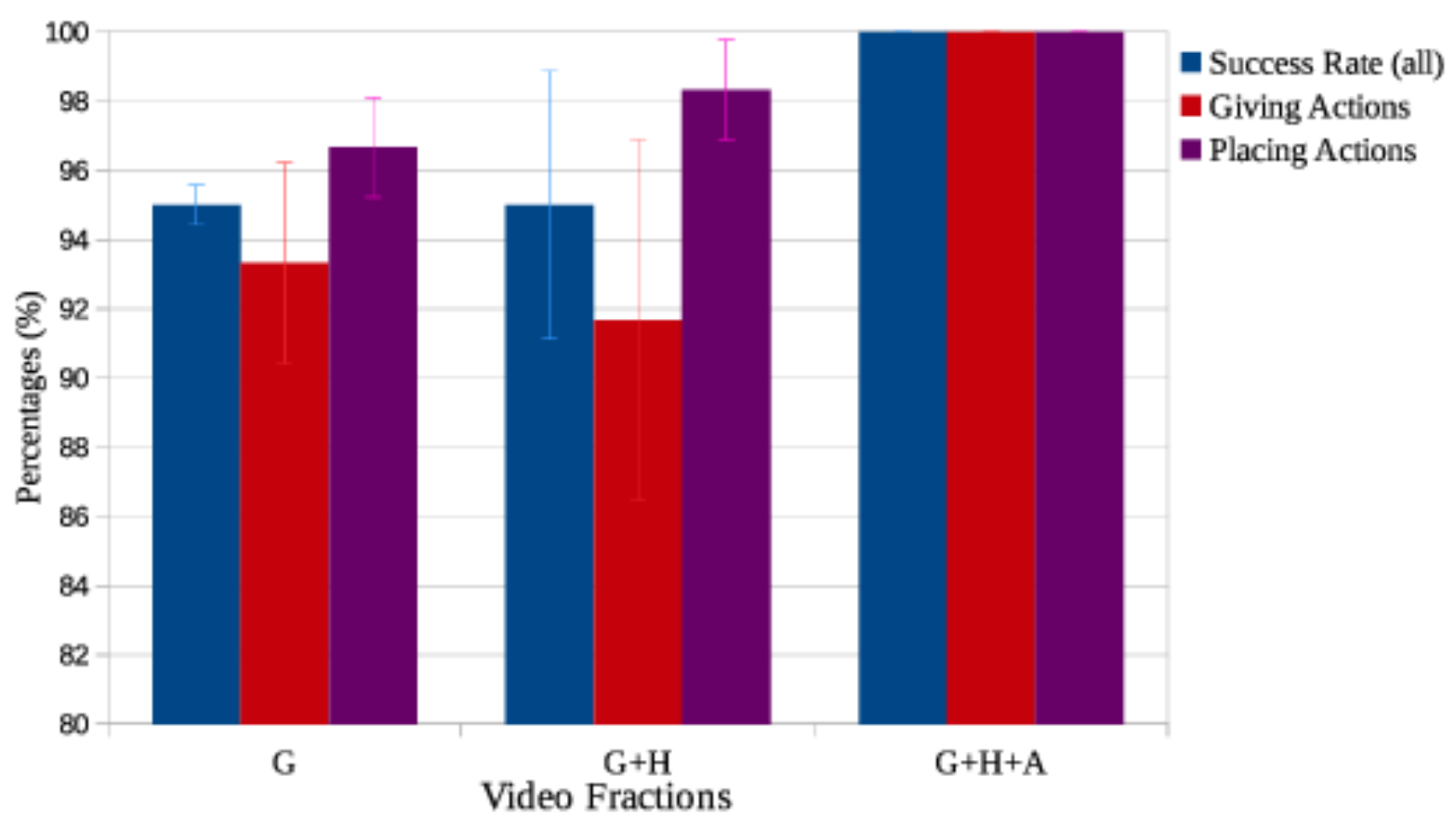}
		\caption{}\label{fig:overall_success_robot}
	\end{subfigure}
	\begin{subfigure}{0.52\textwidth}
		\includegraphics[width = 0.99\textwidth, height=5cm]{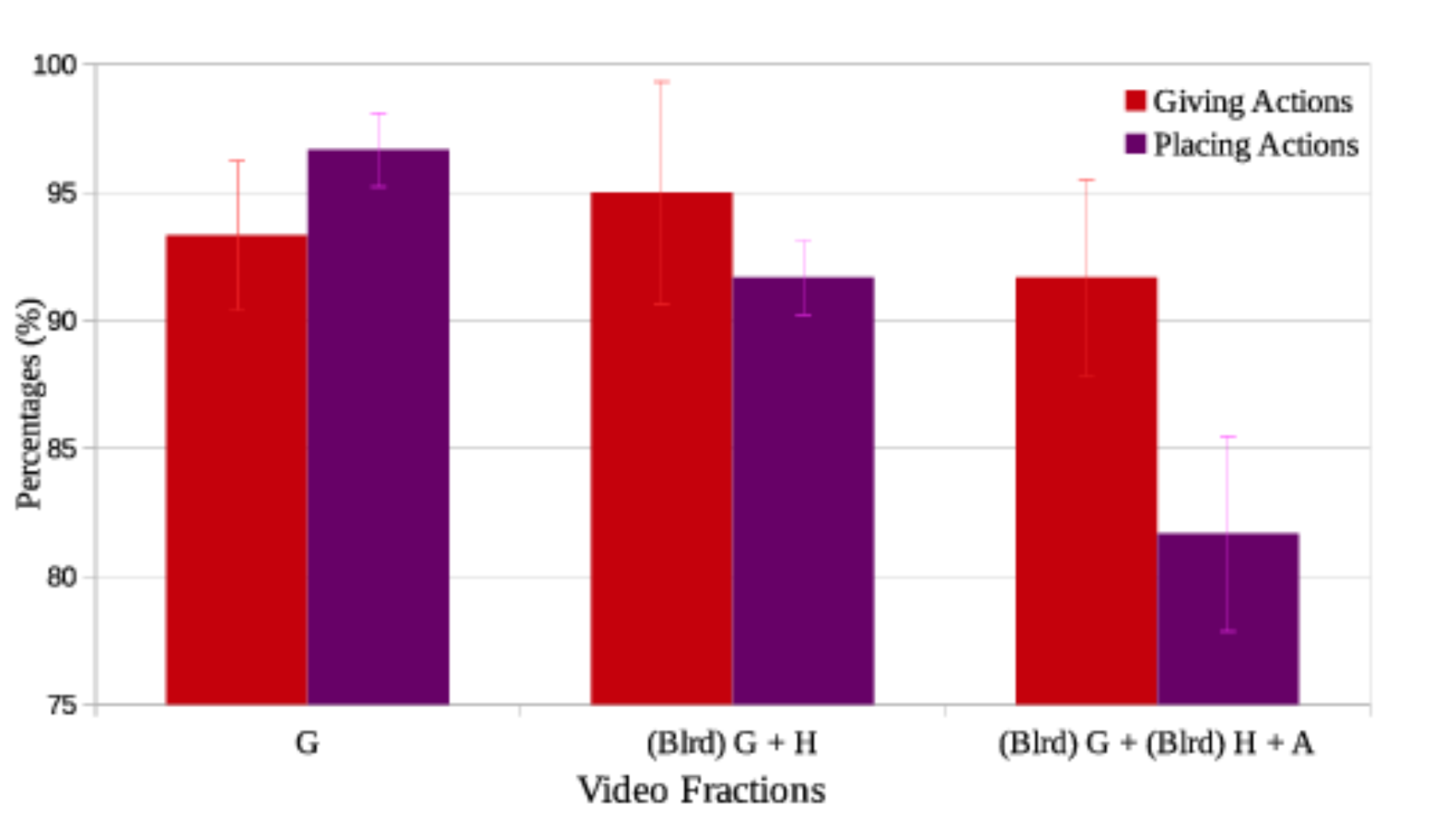}
		\caption{}\label{fig:placing_giving_success_robot}
	\end{subfigure}				
	\caption{The success of the participants identifying the correct action: a) Comparison between the overall success, \textit{giving} actions, and \textit{placing} actions, throughout all the different video fractions; b) The effects of blurring in the success rate. The '(Blrd)' indication after one non-verbal cues, means that one or more non-verbal cues were blurred, in those video fractions. For example, '(Blrd) G + H' means that the eyes were blurred in those video fractions, thus preventing the use of gaze information to read the robot's intention. Nevertheless, in those videos the orientation of the head was still visible. The y axis starts from 80$\%$ for visualization purposes. Gaze and head information (G+H) is in (a), success rate for \textit{giving} or \textit{placing} actions for the blurred gaze and just head information (blurred G + H) is in (b)}
	\label{fig:plots_human_human_robot}
\end{figure}

This second human study involved 20 participants answering 36 questions: 18 without any blurring; 12 with eye blurring, and 6 with the whole head blurred. The 12 eyes blurred questions correspond to the 6$\times$2 possible action-configurations with: (i) blurred eye gaze and with visible head gaze and (ii) blurred eye gaze, with visible head gaze, and visible arm movement); the 6 whole head blurred questions correspond to the 6$\times$1 possible action-configurations with blurred eye gaze, blurred head gaze, and visible arm movement). There were less participants in this second study but each subject had to answer to more questions than in the previous case. Fig.~\ref{fig:overall_success_robot} shows the participants success rate in identifying the robot-action in the three cases: \textit{giving} action, \textit{placing} action or both. 

As in the first study, we observed that the more temporal information the subjects had, the better their decision was, and the higher the success rate. The average success rate increases as more information is provided, Fig.~\ref{fig:plots_human_human_robot}(a).  

In addition, we  analyse the effects of blurring on the success rate of \textit{placing} and \textit{giving} actions, Fig. \ref{fig:plots_human_human_robot}(b). We can see that when blurring the eyes, and preserving only the head information ('(Blrd) G+H') the success rate drops around 5\%. Since there is a clear distinction between the head orientation in \textit{placing} and \textit{giving} actions, for most people, this is enough information to predict the robot's intention. When blurring the whole head, the only information available is the motion of the arm. In \cite{dragan2013legibility}, this motion is classified as "predictable", as such, it will not give the most information to the user. 

Our experiment showed that the difficulty increases with the increase of the blurred area. This means that the legibility of the robot's actions improves with the integration of human-like eye gaze behaviour into the controller. Our work generalizes Dragan et al. \cite{dragan2013legibility}, as legibility is achieved through the combination of both human arm, body, and eye-gaze movements. 

Our results show the importance of non-verbal cues in a human-human interaction scenario, and we successfully transferred the models to a human-robot experiment, where human-level action-readability of robot actions was achieved.

\section{Discussions and Conclusion}\label{sec:discussion}

We conducted experiments and studies to investigate the human ability to read the intention of a human actor during \textit{placing} and \textit{giving} actions in 6 pre-selected action-configurations. One of our contributions was a publicly available dataset of synchronized videos, gaze and body motion data. We conducted an HHI experiment with two objectives: (i) understand how the participants manage to predict the observed actions of the actor; (ii) use the collected data to model the human arm behaviour and (in a qualitative sense) the eye gaze behaviour. 
 
With the human study data, we analysed the different types of non-verbal cues during interpersonal interaction: eye gaze, head gaze, and arm information. For the \textit{placing} actions, with just eye gaze information 85$\%$ of subjects can read the intentions of the actor correctly (chance 50\%). However, for the \textit{giving} actions the results were much worse (chance 20 $\%$). To understand the reason behind these results we analysed the eye gaze behaviour recordings from the eye-tracking system for the \textit{giving} actions. The analysis of these data shows that for the same type of action, there are different gaze trajectories Fig.~\ref{fig:gaze_seq} b-e. According to Moon et al. \cite{moon2014meet}, humans prefer a \textit{giving} action when the actor performs this switching behaviour \cite{zheng2015impacts} observed in Fig.~\ref{fig:gaze_seq} d-e. This switching behaviour can be seen as a confirmation routine to acknowledge to the other person that an interaction is taking place. Since the human motion is a combination of eyes, head, and arm movement, coupled during the action execution, the ``communication'' is only properly established, once it is signalled with this behaviour. As such, the logical choice is to infer that the actor is not trying to communicate with us, which justifies the preference for the \textit{placing} action.

After the analysis, our next contribution was on modelling the human behaviour from the data collected. The arm movements of the actor were modelled with GMM/GMR that can replicate the natural movement of the human arm. Dragan et. al. \cite{dragan2013legibility} proposed two types of arm movements (predictable and legible), and demonstrated that a legible arm movement, which is an overemphasised predictable motion of the human arm, can give more information about the action that the human or the robot is going to do. The experimental scenario involved two end-goals, close to each other. The participants were faster and more accurate to predict the end goal in the case of the overemphasised arm movement. However, there were only very few options in that scenario, and we argue that it would not generalize well if there were more end-goals (for example six as in our case).

We propose an alternative to embed action legibility with overemphasized arm motions, and extend the motion model to incorporate eye gaze information. Our approach improves legibility, by coordinating human-like eye-gaze behaviour with natural arm movements. The resulting robot's behaviour showed to be legible even for multiple sets of actions.  

We validated these findings with a second human study, where subjects had to read/predict the intentions of a robot. In our experiments, it was much easier to read intentions of a robot than those of a human. We can explain this by looking at Fig.\ref{fig:rob_seq}, that shows a side by side comparison of the action performed by the human and the robot. In the second pair of images, we see already a clear change in the eyes of the iCub, which is not yet visible in the case of the human actor. This can be due to the high contrast between the white face and black eyes of the iCub. A different perspective on these results will be addressed in discussion of future work. A link for the video is provided here to illustrate the different steps taken in this work - \href{https://youtu.be/HirRPgZGgFA}{video.ACTICIPATE.ral-2018}.

The final conclusion taken from the second human study is the importance of the robot's gaze for the overall readability of the coordinated motion. Fig.\ref{fig:plots_human_human_robot} shows that just by looking at the arms without any gaze information the success rate drops below 85$\%$. This also results in a slower prediction since the subjects have to wait for the arm of the robot to start moving which is slower than the movement of the eyes. Although 85$\%$ is a good result, it is only when we combine eyes and head movement that the results reach an almost perfect score. Our proposal combines the  human gaze behaviour with the human arm movement to achieve legible behaviour to humans.  

In the future we plan to improve our work in several ways, e.g. by expanding our dataset to more actors. We plan to revisit the modelling of the arm in order to better coordinate the overall eyes/head/arm speed. In our implementation, the robot arm controller is slower than the actual human arm motion. Moreover, while we carefully modelled the arm trajectories using GMMs, the gaze switching behaviour was not modelled with the same level of detail. While the robot gaze controller could qualitatively reproduce the human gaze-shift behaviours, ``the human likeness" were not so close. We will thus investigate methodologies to model the gaze shift dynamics to a greater detail.

Our work stems the importance of non-verbal cues during a HHI, and the benefit of affording robots with the two-fold capacity: (i) interpreting those cues to read the action intentions of their human counterparts and (ii) to act in a way that is legible and predictable to humans.

\balance


%%%%%%%%%%%%%%%%%%%%%%%%%%%%%%%%%%%%%%%%%%%%%%%%%%%%%%%%%%%%%%%%%%%%%%%%%%%%%%%%%
%\section*{APPENDIX}
%\section*{ACKNOWLEDGMENT}
%
%The preferred spelling of the word �acknowledgment� in America is without an �e� after the �g�. Avoid the stilted expression, �One of us (R. B. G.) thanks . . .�  Instead, try �R. B. G. thanks�. Put sponsor acknowledgments in the unnumbered footnote on the first page.

%%%%%%%%%%%%%%%%%%%%%%%%%%%%%%%%%%%%%%%%%%%%%%%%%%%%%%%%%%%%%%%%%%%%%%%%%%%%%%%%
\bibliographystyle{ieeetr}
\bibliography{IEEEabrv,root}

\begin{thebibliography}{10}

\bibitem{dragan2013legibility}
A.~D. Dragan, K.~C.~T. Lee, and S.~S. Srinivasa, ``Legibility and
  predictability of robot motion,'' in {\em 2013 8th ACM/IEEE International
  Conference on Human-Robot Interaction (HRI)}, pp.~301--308, March.

\bibitem{erlhagen2006goal}
W.~Erlhagen, A.~Mukovskiy, E.~Bicho, G.~Panin, C.~Kiss, A.~Knoll, H.~van Schie,
  and H.~Bekkering, ``Goal-directed imitation for robots: A bio-inspired
  approach to action understanding and skill learning,'' {\em Robotics and
  Autonomous Systems}, vol.~54, no.~5, pp.~353 -- 360, 2006.
\newblock The Social Mechanisms of Robot Programming from Demonstration.

\bibitem{huber2008human}
M.~Huber, M.~Rickert, A.~Knoll, T.~Brandt, and S.~Glasauer, ``Human-robot
  interaction in handing-over tasks,'' in {\em RO-MAN 2008 - The 17th IEEE
  International Symposium on Robot and Human Interactive Communication},
  pp.~107--112, Aug 2008.

\bibitem{marin2009interpersonal}
L.~Marin, J.~Issartel, and T.~Chaminade, ``Interpersonal motor coordination:
  From human-human to human-robot interactions,'' vol.~10, pp.~479--504, 12
  2009.

\bibitem{sciutti2018humanizing}
A.~Sciutti, M.~Mara, V.~Tagliasco, and G.~Sandini, ``Humanizing human-robot
  interaction: On the importance of mutual understanding,'' {\em IEEE
  Technology and Society Magazine}, vol.~37, no.~1, pp.~22--29, 2018.

\bibitem{activity2012forecasting}
K.~M. Kitani, B.~D. Ziebart, J.~A. Bagnell, and M.~Hebert, ``Activity
  forecasting,'' in {\em Computer Vision -- ECCV 2012} (A.~Fitzgibbon,
  S.~Lazebnik, P.~Perona, Y.~Sato, and C.~Schmid, eds.), (Berlin, Heidelberg),
  pp.~201--214, Springer Berlin Heidelberg, 2012.

\bibitem{pfeiffer2016predicting}
M.~Pfeiffer, U.~Schwesinger, H.~Sommer, E.~Galceran, and R.~Siegwart,
  ``Predicting actions to act predictably: Cooperative partial motion planning
  with maximum entropy models,'' in {\em Intelligent Robots and Systems (IROS),
  2016 IEEE/RSJ International Conference on}, pp.~2096--2101, IEEE, 2016.

\bibitem{sciutti2015investigating}
A.~Sciutti, C.~Ansuini, C.~Becchio, and G.~Sandini, ``Investigating the ability
  to read others’ intentions using humanoid robots,'' {\em Frontiers in
  psychology}, vol.~6, p.~1362, 2015.

\bibitem{zhang2016rgb}
J.~Zhang, W.~Li, P.~O. Ogunbona, P.~Wang, and C.~Tang, ``Rgb-d-based action
  recognition datasets: A survey,'' {\em Pattern Recognition}, vol.~60, pp.~86
  -- 105, 2016.

\bibitem{koppula2016anticipating}
H.~S. Koppula and A.~Saxena, ``Anticipating human activities using object
  affordances for reactive robotic response,'' {\em IEEE Transactions on
  Pattern Analysis and Machine Intelligence}, vol.~38, pp.~14--29, Jan 2016.

\bibitem{gottwald2015good}
J.~M. Gottwald, B.~Elsner, and O.~Pollatos, ``Good is upâspatial metaphors
  in action observation,'' {\em Frontiers in Psychology}, vol.~6, p.~1605,
  2015.

\bibitem{admoni2014deliberate}
H.~Admoni, A.~Dragan, S.~S. Srinivasa, and B.~Scassellati, ``Deliberate delays
  during robot-to-human handovers improve compliance with gaze communication,''
  in {\em Proceedings of the 2014 ACM/IEEE International Conference on
  Human-robot Interaction}, HRI '14, (New York, NY, USA), pp.~49--56, ACM,
  2014.

\bibitem{fathi2011learning}
A.~Fathi, X.~Ren, and J.~M. Rehg, ``Learning to recognize objects in egocentric
  activities,'' in {\em Proceedings of the 2011 IEEE Conference on Computer
  Vision and Pattern Recognition}, CVPR '11, (Washington, DC, USA),
  pp.~3281--3288, IEEE Computer Society, 2011.

\bibitem{schydlo2018}
P.~Schydlo, M.~Rakovi\'{c}, and J.~Santos-Victor, ``Anticipation in human-robot
  cooperation: A recurrent neural network approach for multiple action
  sequences prediction,'' in {\em accepted for publication at Robotics and
  Automation (ICRA), IEEE International Conference on}, IEEE, 2018.

\bibitem{goodale2011transforming}
M.~A. Goodale, ``Transforming vision into action,'' {\em Vision Research},
  vol.~51, no.~13, pp.~1567 -- 1587, 2011.
\newblock Vision Research 50th Anniversary Issue: Part 2.

\bibitem{ohyama2003cerebellum}
T.~Ohyama, W.~L. Nores, M.~Murphy, and M.~D. Mauk, ``What the cerebellum
  computes,'' {\em Trends in neurosciences}, vol.~26, no.~4, pp.~222--227,
  2003.

\bibitem{shukla2012coupled}
A.~Shukla and A.~Billard, ``Coupled dynamical system based armâhand
  grasping model for learning fast adaptation strategies,'' {\em Robotics and
  Autonomous Systems}, vol.~60, no.~3, pp.~424--440, 2012.
\newblock Autonomous Grasping.

\bibitem{lukic2015visuomotor9}
L.~Lukic, ``Visuomotor coordination in reach-to-grasp tasks: From humans to
  humanoids and vice versa,'' p.~141, 2015.
\newblock Co-supervision with: Instituto Superior TÃ©cnico (IST) da
  Universidade de Lisboa.

\bibitem{elsner2014infants}
C.~Elsner, M.~Bakker, K.~Rohlfing, and G.~GredebÃ€ck, ``Infantsâ online
  perception of give-and-take interactions,'' {\em Journal of Experimental
  Child Psychology}, vol.~126, pp.~280 -- 294, 2014.

\bibitem{land1999roles}
M.~Land, N.~Mennie, and J.~Rusted, ``The roles of vision and eye movements in
  the control of activities of daily living,'' {\em Perception}, vol.~28,
  no.~11, pp.~1311--1328, 1999.
\newblock PMID: 10755142.

\bibitem{zheng2015impacts}
M.~Zheng, A.~Moon, E.~A. Croft, and M.~Q.-H. Meng, ``Impacts of robot head gaze
  on robot-to-human handovers,'' {\em International Journal of Social
  Robotics}, vol.~7, pp.~783--798, Nov 2015.

\bibitem{PrezDArpino2015FastTP}
C.~P{\'e}rez-D'Arpino and J.~A. Shah, ``Fast target prediction of human
  reaching motion for cooperative human-robot manipulation tasks using time
  series classification,'' {\em 2015 IEEE International Conference on Robotics
  and Automation (ICRA)}, pp.~6175--6182, 2015.

\bibitem{hrcollaborative2015julieshah}
S.~Nikolaidis, R.~Ramakrishnan, K.~Gu, and J.~Shah, ``Efficient model learning
  from joint-action demonstrations for human-robot collaborative tasks,'' in
  {\em Proceedings of the Tenth Annual ACM/IEEE International Conference on
  Human-Robot Interaction}, HRI '15, (New York, NY, USA), pp.~189--196, ACM,
  2015.

\bibitem{palinko2016robot}
O.~Palinko, F.~Rea, G.~Sandini, and A.~Sciutti, ``Robot reading human gaze: Why
  eye tracking is better than head tracking for human-robot collaboration,'' in
  {\em 2016 IEEE/RSJ International Conference on Intelligent Robots and Systems
  (IROS)}, pp.~5048--5054, Oct 2016.

\bibitem{Kassner:2014:POS:2638728.2641695}
M.~Kassner, W.~Patera, and A.~Bulling, ``Pupil: An open source platform for
  pervasive eye tracking and mobile gaze-based interaction,'' in {\em Adjunct
  Proceedings of the 2014 ACM International Joint Conference on Pervasive and
  Ubiquitous Computing}, UbiComp '14 Adjunct, (New York, NY, USA),
  pp.~1151--1160, ACM, 2014.

\bibitem{kothe2018lab}
C.~Kothe, ``Lab streaming layer (lsl),'' {\em https://github.
  com/sccn/labstreaminglayer. Accessed on February}.

\bibitem{montgomery2010applied}
D.~C. Montgomery and G.~C. Runger, {\em Applied statistics and probability for
  engineers}.
\newblock John Wiley \& Sons, 2010.

\bibitem{Calinon07}
S.~Calinon, F.~Guenter, and A.~Billard, ``On learning, representing and
  generalizing a task in a humanoid robot,'' {\em IEEE Transactions on Systems,
  Man and Cybernetics, Part B}, vol.~37, no.~2, pp.~286--298, 2007.

\bibitem{moon2014meet}
A.~Moon, D.~M. Troniak, B.~Gleeson, M.~K. Pan, M.~Zheng, B.~A. Blumer,
  K.~MacLean, and E.~A. Croft, ``Meet me where i'm gazing: How shared attention
  gaze affects human-robot handover timing,'' in {\em Proceedings of the
  ACM/IEEE International Conference on Human-robot Interaction}, (New York, NY,
  USA), pp.~334--341, ACM, 2014.

\bibitem{roncone2016cartesian}
A.~Roncone, U.~Pattacini, G.~Metta, and L.~Natale, ``A cartesian 6-dof gaze
  controller for humanoid robots,'' in {\em Robotics: Science and Systems},
  2016.

\end{thebibliography}

\end{document}